\documentclass[a4paper,11pt]{article}
\usepackage{amsfonts}
\usepackage{amssymb}
\usepackage{latexsym}
\usepackage{enumerate}
\usepackage{amsmath}
\usepackage{amsthm}
\usepackage{graphicx}
\usepackage{multirow}
\usepackage{epstopdf}
\usepackage{epigraph} 
\usepackage{caption}
\usepackage{pdflscape}
\usepackage{standalone}
\usepackage{booktabs}
\usepackage{tabularx}
\usepackage[margin=1in]{geometry}
\DeclareGraphicsExtensions{.eps .pdf,.png,.jpg,.mps,.bmp}
\renewcommand{\baselinestretch}{1.2}

\usepackage{verbatimbox}

\begin{document}

\title{Optimal Policy Learning with Observational Data in Multi-Action Scenarios: \\ Estimation, Risk Preference, and Potential Failures} 

\author{Giovanni Cerulli\footnote{This work was supported by the project FOSSR (Fostering Open Science in Social Science Research), funded by the European Union - NextGenerationEU under NPRR Grant agreement n. MUR IR0000008.
The content of this article reflects only the author's view. The European Commission and MUR are not responsible for any use that may be made of the information it contains. The author declares no conflicts of interest regarding this work.} \\ {\small CNR-IRCRES} \\ {\small Research Institute on Sustainable Economic Growth} \\ 
{\small National Research Council of Italy} \\ {\small Via dei Taurini 19, 00185 Rome, Italy} \\  {\small giovanni.cerulli@ircres.cnr.it} \\ {\small \textbf{\textit{}}}}

\date{}

\maketitle

\begin{abstract}
This paper deals with optimal policy learning (OPL) with observational data, i.e. data-driven optimal decision-making, in multi-action (or multi-arm) settings, where a finite set of decision options is available. It is organized in three parts, where I discuss respectively: estimation, risk preference, and potential failures. 
The first part provides a brief review of the key approaches to estimating the reward (or value) function and optimal policy within this context of analysis. Here, I delineate the identification assumptions and statistical properties related to offline optimal policy learning estimators.
In the second part, I delve into the analysis of decision risk. This analysis reveals that the optimal choice can be influenced by the decision maker's attitude towards risks, specifically in terms of the trade-off between reward conditional mean and conditional variance. Here, I present an application of the proposed model to real data, illustrating that the average regret of a policy with multi-valued treatment is contingent on the decision-maker's attitude towards risk.
The third part of the paper discusses the limitations of optimal data-driven decision-making by highlighting conditions under which decision-making can falter. This aspect is linked to the failure of the two fundamental assumptions essential for identifying the optimal choice: (i) \textit{overlapping}, and (ii) \textit{unconfoundedness}. Some conclusions end the paper.

\end{abstract}
\renewcommand{\baselinestretch}{1}
%

\newpage

\section{Introduction}
Decision-making over finite alternatives is a common problem in many domains, ranging from finance to medicine to marketing. The problem of finite-alternative decision-making involves selecting one of several possible options based on a set of input variables (or features) with the goal of maximizing a given reward (or outcome). In the literature, this optimizing procedure is known as \textit{optimal policy learning} (OPL), where the policy is a decision rule mapping a specific configuration of the features (loosely representing the \textit{context} or \textit{environment}) onto a specific action/decision to undertake. This framework is general, and has applications in diverse domains. 

In medicine, for example, personalized medical treatment involves tailoring medical interventions to the unique characteristics of individual patients. This approach recognizes that people differ not only in their health conditions but also in their genetic makeup, lifestyle, and other unique factors. In this case, actions can take form of drugs, surgeries, or alternative therapies to be offered to the patients with the aim of maximizing, for example, the timing of recovery from a given disease. 

In digital advertising, customized product recommendations involve personalized suggestions for products or services that are presented to users based on their preferences, behavior, or historical interactions with the web. This process aims to achieve an optimal allocation of ads with the goal of maximizing sales or future profits.

In the domain of finance, especially within the framework of brokerage and stock trading, a multi-action setting can arise in relation to the process of deciding to purchase one specific stock from a range of available options, with the aim of maximizing capital gains. This involves a meticulous evaluation of diverse factors, including past stock performance, market conditions, and other idiosyncratic elements. 

In the realm of public policies, governments may be responsible for determining the distribution of various forms of financial support to companies based on their individual characteristics. This could involve allocating grants, providing favorable loans, or offering tax credits in a manner that is tailored to each company's unique attributes. The overarching goal might be that of maximizing future companies' financial soundness. The allocation of these resources may be done with the intention of fostering economic growth and success for beneficiary businesses.

In all these contexts of application, data-driven  machine learning algorithms can be applied to automate the decision-making process, as they can learn from past (observed) data and make predictions about which alternative is most likely to maximize the reward (Marabelli et al. 2021; Xin et al., 2020; Wen and Li, 2023). The use of OPL for data-driven decision-making has proved to lead to faster and more accurate decisions, as well as more efficient allocation of resources, compared to qualitative approaches or to approaches based on descriptive or anecdotal evidence (Tschernutter, 2022). 

This paper considers data structured as a triplet: (i) a signal from the environment, comprising a series of observed features; (ii) a set of multiple actions from which one is chosen; and (iii) a reward associated with the selected action. This data structure accommodates two distinct scenarios.

The first scenario pertains to the behavior of a single agent attempting to maximize a specified reward while performing a particular task. For instance, a company may have accumulated data over time regarding its operational context (market conditions, competitors' prices, previous sales, etc.), the types of advertisements utilized (web ads, TV commercials, newspaper ads), and the resulting sales. Given a new environmental signal, the company can leverage this information to formulate an advertising strategy that maximizes sales. Consequently, the data pertains to the same company, representing a context that can be described as \textit{agent-based}, with the company playing the role of the agent. This scenario fits well also robotics applications, where a robot can exploit observational data to learn, for instance, how to reach a certain place or how to move a given object. In this case, OPL with observational data can be encompassed within the so-called \textit{imitation learning}, where data are made of a collection of \textit{context-action-reward} triplets previously experienced by the robot itself, humans, or even other robots (Zheng et al., 2021; Hussein et al., 2017).  

The second scenario involves collecting data triplets from different agents who have taken diverse actions in response to distinct environmental signals experienced in the past. For instance, the data could include information on multiple patients arriving at an emergency room, requiring a doctor to assess their health status as ``good'', ``very good'', ``bad'', or ``severely bad'' to prioritize cases with more compromised health conditions. In this context, OPL involves evaluating the health conditions of individuals to optimally allocate them to a specific health status with aim of reducing as much as possible potential mis-classifications. Similarly, a social planner might determine which unemployed individuals should or should not receive specific social support based on previously gathered characteristics of these individuals and an observed reward, such as employment status (employed vs. unemployed) some time after the provision of the support.         

In the second scenario, it is essential to operate under the assumption that observations are independent and identically distributed (i.i.d.). This assumption, however, cannot be maintained in the first scenario due to the inherent path-dependence characterizing decisions. Nevertheless, in this case, the i.i.d. assumption can still be applied if conditional on past decisions (as time matters in this case). This paper assumes as reference the second scenario, but many results can be easily generalized also to the first scenario with only minor changes. 

With proper adjustments, both decision settings can be encompassed within the so-called \textit{contextual multi-armed bandit with observational data}, a simple yet powerful framework used in machine learning and decision-making problems to select optimal actions using data (Auer et al. 2002; Slivkins, 2019; Silva et al., 2022). 

As part of the branch of machine learning called \textit{reinforcement learning} (Sutton and Barto, 2018; Li, 2023; ), the name ``bandit'' comes from the idea of a slot machine, where each arm corresponds to a lever that can be pulled, and the rewards are payouts. The term ``multi-armed'' indicates that there are multiple levers to pull, each with its own payout probability (Sutton \& Barto, 1998; Silva et al., 2022; Bouneffouf et al., 2020; Mui and Dewan, 2021).

In the canonical multi-armed bandit, actions' reward probabilities are unknown, and the goal is to find the optimal arm (or action) that maximizes the cumulative reward over a certain number of rounds, or minimizes the so called \textit{regret} defined as the difference between the average cumulative reward that the agent would obtain if she was pulling at each round the best arm, and the average cumulative reward of the options actually chosen at each round.  

As the agent does not initially know the reward probabilities of each arm, she must explore the different options to learn more about them while simultaneously exploiting the best arm currently found. This leads to the emergence of a trade-off between \textit{exploration} and \textit{exploitation}: wider exploration increases the chance to discover more rewarding actions, but prevent at the same time to exploit those options that have been proved to be -- so far -- more rewarding; on the contrary, deeper exploitation allows for obtaining higher rewards from the options that have been proved to be more rewarding, but can run the risk to let the agent stuck to a sub-optimal solution.   

The literature has proposed several algorithms to solve the contextual multi-armed bandit, where -- by solution -- they intend an algorithm able to detect -- after a certain number of steps -- the arm with the largest average reward\footnote{One common approach to solving the multi-armed bandit problem is called the \textit{epsilon-greedy} algorithm (Kuang \& Leung, 2019; Rawson \& Balan, 2021). In this algorithm, the agent selects the arm with the highest estimated reward with probability (1 - \textit{epsilon}), and selects a random arm with probability \textit{epsilon}. This approach balances exploration and exploitation by encouraging the agent to occasionally choose a less-known arm to gather more information.
Another approach is the \textit{upper confidence bound} (UCB) algorithm (Takeno et al., 2023; Rawson and Freeman, 2021; Zhu et al., 2021). This algorithm selects the arm with the highest upper confidence bound, which is a measure of how uncertain the agent is about the reward probability of each arm. The UCB algorithm tends to be more efficient than epsilon-greedy in situations where the rewards are sparse or non-stationary.} (Agarwal et al., 2014).  

OPL with observational data starts by assuming that it already exists a sufficiently extensive set of available information collected over the past. If this dataset includes environmental signals, actions taken, and corresponding rewards, the exploration phase needed to recover the reward probability of each arm can be bypassed. Indeed, one can directly discover the decision rule that selects the best action using a pure exploitative (data-driven) approach. This process relies on maximizing the empirical reward, subject to specific assumptions about the statistical identification of the best choice (more later on).

Two different modes of learning are generally used in OPL with observational data: \textit{offline} and \textit{online} learning. In the offline, the entire dataset is available from the start, while online learning handles data that arrives sequentially (typically over time). Offline learning updates the model's parameters after processing the entire dataset, whereas online learning updates the model incrementally as new an instance arrives. Offline learning is suitable for static, medium-sized datasets, where refitting the model to the data as new an instance arrives does not involve severe computational burden. On the contrary, online learning is suitable in contexts characterized by dynamic, streaming, or rapidly changing Big Data (billions of observations), where the computational cost of refitting the learning model over the entire dataset would be prohibitive. Although more focused on offline learning, this study also discusses online OPL.

This paper is organized in three parts: The first part provides a brief review of the key approaches to estimating the reward (or value) function and optimal policy within offline OPL with observational data. Here, I delineate the identification assumptions and statistical properties related to the main offline optimal policy learning estimators provided by the literature.

In the second part, with a focus on online learning, I delve into the analysis of decision risk. This analysis reveals that the optimal choice can be influenced by the decision maker's willingness to take risks, specifically in terms of the trade-off between reward conditional mean and conditional variance. This demonstrates that a purely objective, data-driven approach to optimal decision-making (i.e., OPL) is not feasible. Here, I present an application of the proposed model to real data, illustrating that the regret of the policy is contingent on the decision-maker's attitude towards risk.

The third part of the paper discusses the limitations of data-driven OPL, by highlighting conditions under which decision-making can falter. This aspect is linked to the failure of the two fundamental assumptions essential for identifying the optimal choice: (i) unconfoundedness, and (ii) overlapping. Some conclusions end the paper.

\section{Offline optimal policy learning}

Consider a set of $N$ observations indexed by $i=1,\dots,N$, and a set of $J+1$ different actions/decisions $D_{i}=0,1,2,\dots, j, \dots ,J$. Associated to each action/decision, we define a set of $J+1$ potential rewards $\{Y_{i}(0),Y_{i}(1),\dots,Y_{i}(J)\}$ having statistical distributions $\{\mathcal{F}_{i}(0), \mathcal{F}_{i}(1),\dots,\mathcal{F}_{i}(J)\}$. For each observation, we also define a vector of $p$ predictors (or features) $\textbf{x}_{i}$.

In the context of policy learning, a policy is defined as a function mapping  $\textbf{x}$ onto $j$, i.e.:
\begin{equation} \label{eq:policy1}
\pi: \text{  } \textbf{x} \longrightarrow j \in \{0, 1, \dots , J+1 \}
\end{equation}
implying that:
\begin{equation} \label{eq:policy2}
j=\pi(\textbf{x}).
\end{equation}
Associated to a given policy $\pi$, we define the \textit{value function} as:
\begin{equation} \label{eq:valfun}
V(\pi) = \text{E}[Y(\pi(\textbf{x})]
\end{equation}
which is a scalar indicating the welfare achieved by policy $\pi$. 
An optimal policy $\pi^{*}$, within a class of policies $\Pi$, is defined as:
\begin{equation} \label{eq:valfun2}
\pi^{*} = \text{argmax}_{\pi \in \Pi} \text{E}[Y(\pi)]
\end{equation}
For a given policy $\pi \in \Pi$, we define the so-called \textit{regret} as:

\begin{equation} \label{eq:regret}
R(\pi)= \text{E}[Y(\pi^{*})] - \text{E}[Y(\pi)] = V(\pi^{*}) - V(\pi)
\end{equation}
which identifies a loss of welfare, whenever $Y$ indicates a measure of welfare (as, for example, personal income).

A fundamental task of policy learning is to estimate the optimal policy $\pi^{*} \in \Pi$ and the corresponding value function (i.e., the average reward) starting from observing $N$ independent and identically distributed observations $\{(\textbf{x}_{i}, D_{i}, Y_{i})\}_{i=1}^{N}$, where $Y_{i}$ is an observed measure of welfare.   

For this purpose, define the \textit{conditional expected reward} of the observation $i$ when action/decision $j$ is selected as:
\begin{equation} \label{eq:cer}
\mu_{i}(j,\textbf{x}_{i}) = \text{E}(Y_{i}(j)|\textbf{x}_{i})
\end{equation}
In a binary setting, with only two actions/decisions (i.e. $j \in \{0,1\}$), Kitagawa and Tetenov (2018) define the \textit{first-best} optimal rule as:

\begin{equation} \label{eq:fb1}
\pi^{fb}_{i}(\textbf{x}_{i}) = 1[ \mu_{i}(1,\textbf{x}_{i}) \geq \mu_{i}(0,\textbf{x}_{i})]
\end{equation} 
where the indicator function $1[A]$ takes value 1 if $A$ is true and 0 otherwise. 
The policy rule (\ref{eq:fb1})  maximizes the value function (or population welfare) of equation (\ref{eq:valfun}) if whatever assignment rule is feasible to implement. With $J+1$ actions/decisions, the generalized first-best decision rule is:

\begin{equation} \label{eq:fb2}
\pi^{gfb}_{i}(\textbf{x}_{i}) =  j[ \mu_{i}(j,\textbf{x}_{i}) \geq 
\mu_{i}(k,\textbf{x}_{i}) , \text{  } \forall k=0,\dots , j-1,j+1,\dots,J]
\end{equation}
which is the unconstrained optimal policy rule. In many contexts of application, and particularly in the socio-economic context, however, we generally deal with constrained classes of feasible assignment rules 
incorporating several types of exogenous constraints, which restrict the complexity of feasible treatment assignment rules. This may depend on logistic, legal, ethical, or political restrictions.

One of the problem with equation (\ref{eq:fb2}) is that it is expressed in terms of counterfactuals, thus it cannot be estimated by observation. To provide identification of the counterfactuals, two assumptions are generally invoked: 
\\
\\
\textbf{A1. Unconfoundedness (or selection-on-observables)}. For all $j = 0, 1, . . . , J$, and for all $i=1,\dots,N$: \\
$$Y_{i}(j) \perp D_{i} | \textbf{x}_{i}$$ 
\\
This assumption entails that, conditional on the knowledge of the environment (i.e., the vector $\textbf{x}_{i}$), there is statistical independence between the potential outcome when decision variable $j$ is selected and the decision variable $D_{i}$. In other words, A1 entails conditional randomization of the undertaken choice once the signal from the environment has been received. This assumption rules out the possible existence of other environmental components having an effect on $Y_{i}(j)$ and simultaneously on $D_{i}$ (\textit{hidden confounders}).     
\\
\\
\textbf{A2. Overlapping}. For all $j = 0, 1, . . . , J$, and for all $i=1,\dots,N$: \\
 $$0 < p_{min} < p_{j}(\textbf{x}_{i}) \hspace{0.2cm}\text{with}\hspace{0.2cm} p_{j}(\textbf{x}_{i})  = P(D_{i} = j|\textbf{x}_{i})$$
\\
This assumption assumes that the so-called propensity score for action $j$ -- i.e.. $P(D_{i} = j|\textbf{x}_{i})$ -- must never be exactly equal to zero. If it exists an $\textbf{x}_{i} = \textbf{x}^{*}_{i}$ such that $P(D_{i} = j|\textbf{x}_{i})=0$, this means that the probability to observe action $j$ for a specific configuration of the environment is zero. Consequently, for certain configurations of $\textbf{x}_{i}$, we cannot observe action/decision $j$, thus making it impossible to build a mapping between the observed reward $Y_{i}$ and action/decision $j$ when $\textbf{x}_{i} = \textbf{x}^{*}_{i}$.    

Under assumptions A1 and A2, we can prove that (Imbens \&  Rubin, 2015; Cerulli 2022):

\begin{equation} \label{eq:cmi1}
\mu_{i}(j,\textbf{x}_{i}) =  \text{E}(Y_{i}| D_{i}=j ,\textbf{x}_{i}) 
\end{equation}
implying that the first-best policy can be estimated by observation, that is, using the dataset provided by the triplet $(\textbf{x}_{i}, D_{i}, Y_{i})$.

\noindent 
\rule{\linewidth}{0.4pt} 
\textbf{Example 1.} \textit{OPL with linear reward and threshold-based policy class}. 
\\
Consider a reward function which is linear in the policy, and depends on a parameter $c$ as:
\begin{equation} \label{eq:ex2}
Y = \alpha(c) \cdot \pi(X) + \epsilon
\end{equation}
where $\alpha(c)$ is a continuous function in $c$, and $\epsilon$ is a pure random shock (with zero mean and finite variance) uncorrelated with the random variable $X$. Consider the following \textit{threshold-based} policy rule:
\begin{equation} \label{eq:treshPolicy}
\pi(X) = 1[X<c]
\end{equation}
where $c$ is the constant threshold. This implies that:
\begin{equation} \label{eq:ex3}
Y = \alpha(c) \cdot 1[X<c] + \epsilon
\end{equation}
We can define the average reward as:
\begin{equation} \label{eq:ex4}
\text{E}(Y) = \alpha(c) \cdot \text{E}(1[X<c])= \alpha(c) \cdot \text{Prob}(X<c)=\alpha(c) \cdot F_{X}(c)
\end{equation}
where $F_{X}(c)$ is the c.d.f. of $X$ evaluated at $c$. 
We define the optimal policy as:
\begin{equation} \label{eq:ex5}
\pi^{*}(X) = 1[X<c^{*}]
\end{equation}
where:
\begin{equation} \label{eq:ex6}
c^{*}=\text{argmax}_{c} [ \alpha(c) \cdot F_{X}(c)].
\end{equation}
Since $F_{X}(c)$ is monotonically increasing in $c$, being it a c.d.f., the solution turns out to become:
\begin{equation} \label{eq:ex7}
c^{*}=\text{argmax}_{c} \alpha(c).
\end{equation}
If $\alpha(c)$ is concave in $c$, the solution is trivial. Observe that $\alpha(c)$ can be interpreted, for example, as a net-benefit function.  
\\
\noindent
\rule{\linewidth}{0.4pt} 
\noindent
Under assumptions A1 and A2, and correct functional specification, the literature has provided three types of consistent estimates of the value-function as expressed in equation (\ref{eq:valfun}) for a given policy $\pi(\textbf{x})$: \textit{regression adjustment}, \textit{inverse probability weighting}, and the \textit{doubly-robust} estimators (Dudik, Langford, and Li, 2011).  

\begin{enumerate}
\item  \textit{Regression adjustment (RA)}. This approach estimates the value function using regression estimates of the counterfactual (potential) outcomes. As such, it is also known as the \textit{direct method}. The regression adjustment formula is:  

\begin{equation} \label{eq:regAdj}
\hat{V}_{RA}(\pi) = \frac{1}{N}  \sum_{i=1}^{N}  \hat{\mu}_{i}(\pi(\textbf{x}_{i}),\textbf{x}_{i}) 
\end{equation}
where $\hat{\mu}_{i}(\pi(\textbf{x}_{i}),\textbf{x}_{i})=
\sum_{j=0}^{J} \hat{\mu}_{i}(j,\textbf{x}_{i}) \cdot \pi_{ij}$ with 
$\pi_{ij}=1[\pi_{i}=j]$. The RA approach provides a consistent estimation of the value function provided that the functional form of the regression model is correct. If this is not the case, this approach can be highly biased.  

\item   \textit{Inverse probability weighting (IPW)}. The formula of this estimator of the value-function is: 

\begin{equation} \label{eq:ipw}
\hat{V}_{IPW}(\pi) = \frac{1}{N}  \sum_{i=1}^{N} \frac{1[D_{i}=\pi(\textbf{x}_{i})] Y_{i}} {\hat{p}_{D_{i}}(\textbf{x}_{i})}
\end{equation}
where $\hat{p}_{D_{i}}(\textbf{x}_{i})$ is an estimate of the propensity score. The \textit{IPW} approach does not require an estimation of the mean potential outcomes; rather, it uses directly the values of the observed outcome variable $Y$. Unfortunately, this estimation method is biased when the propensity score functional form is misspecified. Interestingly, when the value function to evaluate is that of the current observed policy $D$, the \textit{IPW} estimator becomes:
\begin{equation} \label{eq:ipw2}
\hat{V}_{IPW}(\pi) = \frac{1}{N}  \sum_{i=1}^{N} \frac{Y_{i}} {\hat{p}_{D_{i}}(\textbf{x}_{i})}
\end{equation}
which is the well-known Horvitz \& Thompson (1952) estimator, used for estimating the total and mean of a pseudo-population in a stratified sample. This makes it clear that the \textit{IPW} estimator accounts for different proportions of observations within the action space. 
   
\item   \textit{Doubly-robust (DR)}. This estimator of the value-function, derived from the optimal influence function, takes on this formula: 
\begin{equation} \label{eq:dr}
\hat{V}_{DR}(\pi) = \frac{1}{N}  \sum_{i=1}^{N}  \Biggr[   \frac{[Y_{i}-\hat{\mu}_{i}(D_{i},\textbf{x}_{i})] \cdot 1[D_{i}=\pi(\textbf{x}_{i})]}{\hat{p}_{D_{i}}(\textbf{x}_{i})}+\hat{\mu}_{i}(\pi(\textbf{x}_{i}),\textbf{x}_{i}) \Biggr]
\end{equation}
Unlike the \textit{RA} and \textit{IPW} approaches, the \textit{DR} does not require for its consistency that both the propensity score and the conditional mean are simultaneously correctly specified. Only one out of the two must be correctly specified, with the other being potentially also mispecified. 
\end{enumerate}

\subsection{Constrained policy learning: an example}
The unconstrained optimal policy implied by equation (\ref{eq:fb2}) cannot be viable or practical when certain policy constraints become binding. These constrains can pertain social, legal, ethical or even political issues that can make the implementation of the first-best policy unfeasible. 

We can thus restrict the search for the optimal policy within a restricted class of policies that can have specific characteristics. A popular policy class within a multi-action policy setting is the \textit{threshold-based}. For a three-class setting, and only one feature $x$, this policy class takes on this form:

\begin{equation} \label{eq:TBpolicy}
\pi_{tb}(x_{i}, c_{1}, c_{2}) = 0 \times 1[x_{i} \leq c_{1}] + 1 \times 1[ c_{1} \leq  x_{i} \leq c_{2}] + 2 \times 1[x > c_{2}]
\end{equation}
\begin{figure}[h!]
\centering
\includegraphics[width=11cm]{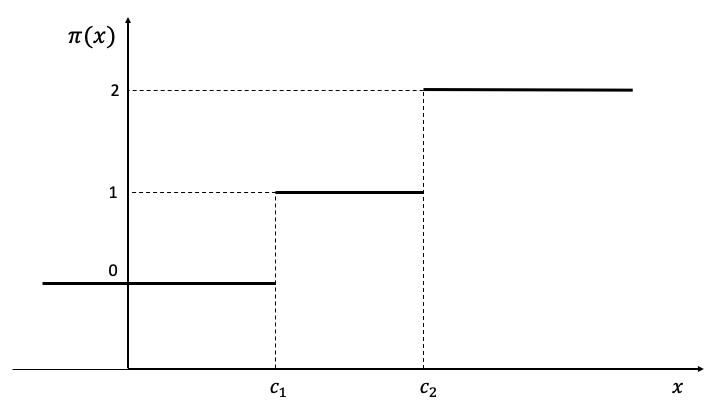}
\caption{Threshold-based policy class.}
\label{fig:thresholdPolicy}
\end{figure}
Figure \ref{fig:thresholdPolicy} draws this policy function which is clearly a step function with knots at $c_{1}$ and $c_{2}$. Finding an optimal policy entails detecting two optimal values for the knots  $c_{1}$ and $c_{2}$.  For example, if we consider the \textit{IPW} estimator of the value-function, the optimal threshold-based policy takes on this form:   

\begin{equation} \label{eq:ipw}
\pi^{*}_{tb}(x_{i}) = \text{argmax}_{(c_{1}, c_{2})} \frac{1}{N}  \sum_{i=1}^{N} \frac{1[D_{i}=\pi_{tb}(x_{i}, c_{1}, c_{2}) ] Y_{i}} {\hat{p}_{D_{i}}(\textbf{x}_{i})}
\end{equation}
with $c_{2} > c_{1}$. The optimal policy can be estimated quite easily computationally by applying Procedure 1 (see below).

\begin{table} [h!]
\centering
\begin{tabular}{>{\hspace{0pt}}m{0.94\linewidth}}  
\rule{\linewidth}{0.4pt} 
\textbf{Procedure 1.} \textit{Computation of the optimal threshold policy}                                                                                                                                                                

\begin{enumerate}

\item Generate a grid of values for the pair $\{c_{1},c_{2}\}$ covering the support of $x$.

\item Generate as many different policies as the ones corresponding to the previously defined grid. 

\item For each policy thus generated, compute the value-function using the \textit{IPW} estimator.  

\item Select the \textit{IPW} estimator having the largest value.
\end{enumerate}
\rule{\linewidth}{0.4pt} 
\end{tabular}
\end{table}
Procedure 1 can also be extended to the \textit{RA} and \textit{DR} estimators of the value function, provided that we utilize their respective formulas in step 3, rather than the \textit{IPW} formula. It's worth noting that in a multi-action scenario, alternative policy classes can be employed. One popular choice is the \textit{fixed-depth tree} policy class, which employs a decision tree to determine the optimal action/decision to take (Zhou, Athey, and Wager, 2023).
   
\subsection{Statistical properties of the value-function estimators}
The purpose of optimal policy learning is to learn a policy, which entails either determining the optimal action an agent should take, or how to allocate treatments among individuals, with the objective of maximizing the value function (or welfare), or alternatively, minimizing the regret. 

It's evident that the accuracy of estimating the value function, and consequently, the optimal policy, hinges on the precision of estimating two key components: the conditional expectation denoted as $\hat{\mu}_{i}(\pi(\textbf{x}_{i}),\textbf{x}_{i})$ and the propensity score denoted as $\hat{p}_{D_{i}}(\textbf{x}_{i})$. When both of these estimates consistently reflect the true conditional expectation and propensity score, both the \textit{RA} and \textit{IPW} estimators yield consistent value-function estimates. However, the \textit{DR} estimator only requires one of these two nuisance parameters to be consistent (not both simultaneously), hence its name \textit{doubly-robust}.

A compelling question arises when we consider how these estimators (\textit{RA}, \textit{IPW}, and \textit{DR}) perform when they deviate from the true value of the value-function. This proves especially valuable for examining the finite sample properties of these estimators, which involves understanding how they behave when the size of the training sample is not very large.

\subsubsection{Computing the \textit{bias}}

Dudik, Langford, and Li (2011) provide bias and variance formulas for the three previous estimators as function of the deviation of $\hat{\mu}_{i}(\pi(\textbf{x}_{i}),\textbf{x}_{i})$ and $\hat{p}_{D_{i}}(\textbf{x}_{i})$ from their true values. For simplicity, call these two quantities as $\hat{\mu}_{\pi}$ and $\hat{p}_{D}$ respectively. Also, define the deviations for both the conditional mean and the propensity score as respectively:
\begin{equation} \label{eq:Delta}
\Delta= \hat{\mu}_{\pi} - \mu_{\pi}
\end{equation} 
and 
\begin{equation} \label{eq:Delta}
\delta = 1 - \frac{p_{D}}{\hat{p}_{D}}
\end{equation} 
It can be proved that the biases of the three estimators are:
\begin{eqnarray}
|\text{E}(V^{\pi}_{RA}) - V^{\pi}| = |\text{E}_{\textbf{x}}(\Delta)|   \\
|\text{E}(V^{\pi}_{IPW}) - V^{\pi}| = |\text{E}_{\textbf{x}}(\mu_{i}(\cdot)\delta)|   \\
|\text{E}(V^{\pi}_{DR}) - V^{\pi}| = |\text{E}_{\textbf{x}}(\Delta \cdot \delta)| 
\end{eqnarray}  
where it is clear that the \textit{DR} estimator has zero bias as long as either $\Delta \approx 0$ or $\delta \approx 0$. On the contrary, the \textit{RA} requires $\Delta \approx 0$, and the \textit{IPW} requires $\delta \approx 0$. In general, in terms of bias, none of the estimators dominates the other. However, when $\Delta \approx 0$ and $\delta \ll 1$, then the \textit{DR} has smaller bias than \textit{RA}, while when $\Delta \gg 0$ and $\delta \approx 0$, the \textit{DR} has smaller bias than the \textit{IPW}.  

\subsubsection{Computing the \textit{variance}}
In terms of variance, it can be proved that:
\begin{eqnarray}
\text{Var}(V^{\pi}_{RA}) = \frac{1}{N} \text{Var}[\mu_{\pi}+ \Delta]  \\
\text{Var}(V^{\pi}_{IPW}) = 
\frac{1}{N} \Biggl( \text{E}[\epsilon_2]+\text{Var}[\mu_{\pi}-\mu_{\pi} \cdot \delta] + \text{E} \Biggr[ \frac{1-p}{p} \cdot \mu_{\pi}^2 (1-\delta)^2 \Biggr]  \Biggl)  \\
\text{Var}(V^{\pi}_{DR}) = 
\frac{1}{N} \Biggl( \text{E}[\epsilon_2]+\text{Var}[\mu_{\pi}+\Delta \cdot \delta] + \text{E}   \Biggr[   \frac{1-p}{p} \cdot \Delta^2 (1-\delta)^2 \Biggr]  \Biggl)  
\end{eqnarray}    
where $p=p_{\pi}$, and $\epsilon=(Y-\mu_{\pi})\cdot 1[\pi_{\textbf{x}}=D]/\hat{p}$.
The variance of the \textit{DR} estimator can be split into three components: one accounting for the randomness in the outcomes; one equal to the variance of the estimator due to the randomness in \textbf{x}, and one reflecting the importance weighting penalty. For the \textit{IPW}, we obtain a similar formula, where the first term is the same as the \textit{DR}, the second term will have similar size of the corresponding term of the \textit{DR} estimator if $\delta \approx 0$, and the third term can be much larger for the \textit{IPW} if $p_{\pi} \ll 1$ and $|\Delta|$ is smaller than $\mu_{\pi}$. The variance of the \textit{RA}, finally, only presents the second term, ensuring that it is remarkably smaller than the variance of the \textit{DR} or \textit{IPW} estimators. Nonetheless, as seen above, the bias of the \textit{RA} is in general much larger than the bias of the \textit{IPW} and \textit{DR}, thus generally providing larger errors in estimating the value-function.

\subsubsection{Rate of convergence}
Even when an estimate of the value-function is consistent, that is, it converges in probability to the true value-function, the rate of convergence seems important to evaluate the quality of the estimator as the sample size $N$ increases: among consistent estimators, faster-to-converge estimators are preferred. 

The recent literature on policy learning using observational data has provided a series of important results concerning the rate of convergence of algorithms mainly based on the \textit{IPW} or \textit{DR} estimators. We start by considering first some relevant results for the binary-action setting:

\begin{itemize}
\item  Zhao et al (2014) developed nonparametric doubly-robust estimator for a censored outcome based on the \textit{IPW} estimator reaching a convergence rate of order $O_{p}(\frac{1}{N^{\frac{1}{2+1/q}}})$, where $q >0$ is a parameter indicating the degree of separation between the two treatment classes. 

\item  Kitagawa and Tetenov (2018) provided an improved \textit{IPW} algorithm reaching a rate of convergence of optimal order $O_{p}(\frac{1}{\sqrt{N}})$, although this rate of convergence requires the knowledge of the underlying propensity score.

\item  Athey and Wager (2021), finally, proposed another \textit{IPW}-based learning algorithm establishing an optimal $O_{p}(\frac{1}{\sqrt{N}})$ regret bound even in the case where the propensity score is unknown and must be estimated.  
\end{itemize}

We consider important results also in the case of a multi-action policy setting:

\begin{itemize}

\item  Swaminathan and Joachims (2015), by addressing the counterfactual nature of the policy learning problem through propensity scoring,  prove a generalization regret bounds that accounts for the variance of the propensity-weighted empirical risk estimator. The proposed Policy Optimizer
for Exponential Models (POEM) provides regret bound converging at speed of order $O_{p}(\frac{1}{N^{1/4}})$. This algorithm requires however a known propensity score. 

\item  Zhou et al. (2017) propose another kind of inverse probability weighting algorithm called Residual Weighted Learning (RWL). Their algorithm, still requires to know the propensity score and provides a rate of converge of the regret of order $O_{p}({N^{-\frac{\beta}{2\beta+1}}})$ (with $0 < \beta \leq 1$) which is  however non-optimal. 

\item  Kallus (2018) have recently proposed methods with formal consistency guarantees for the regret even when the propensity score is unknown and has to be estimated, called the Balanced Policy Learning approach. However the regret bound of Kallus (2018) scales as $O_{p}(\frac{1}{N^{1/4}})$, which is sub-optimal. 
\end{itemize}
So far, the only algorithm reaching asymptotically minimax-optimal regret -- that is, a rate of convergence of the regret with optimal order $O_{p}(\frac{1}{\sqrt{N}})$  --  is the Cross-fitted Augmented Inverse Propensity Weighted Learning (CAIPWL) proposed by Zhou, Athey, and Wager (2023), based on the theory of efficient semi-parametric inference. Given the importance of this algorithm, I provide a schematic account of it. The algorithm entails five steps:

\begin{enumerate}
\item  Consider as input a dataset $\{(\textbf{x}_{i}, D_{i}, Y_{i})\}_{i=1}^{N}$.

\item Split randomly the dataset into $K>1$ folds.  

\item For $k=1,2,\dots,K$: \\

build the estimators: 
$\hat{\mu}^{-k}(\cdot)=
\left(
\begin{array}{c}
\hat{\mu}^{-k}_{0}(\cdot)  \\
\hat{\mu}^{-k}_{1}(\cdot)   \\
\cdots \\
\hat{\mu}^{-k}_{J}(\cdot)  
\end{array}
\right)
$
and
$\hat{p}^{-k}(\cdot)=
\left(
\begin{array}{c}
\hat{p}^{-k}_{0}(\cdot)  \\
\hat{p}^{-k}_{1}(\cdot)   \\
\cdots \\
\hat{p}^{-k}_{J}(\cdot)  
\end{array}
\right)
$
using the remaining $K-1$ folds.

\item Completed the loop over $k$, build the approximate value-function:

$$
\hat{Q}_{CAIPWL}(\pi)= \frac{1}{N} 
\sum_{i=1}^{N} 
\langle  \textbf{d}_{\pi(\textbf{x}_{i})} , 
\frac{Y_{i} -  \hat{\mu}^{-k(i)}_{D_{i}}(\textbf{x}_{i})}{\hat{p}^{-k(i)}_{D_{i}}(\textbf{x}_{i})} \cdot \textbf{d}_{D_{i}} + \left(
\begin{array}{c}
\hat{\mu}^{-k}_{0}(\cdot)  \\
\hat{\mu}^{-k}_{1}(\cdot)   \\
\cdots \\
\hat{\mu}^{-k}_{J}(\cdot)  
\end{array}
\right)
\rangle
$$
where $\langle \cdot , \cdot \rangle$ represents the matrix inner product, $\textbf{d}_{\pi(\textbf{x}_{i})}$ the $J+1$-dimensional basis vector for the policy $\pi(\textbf{x}_{i})$, and $\textbf{d}_{D_{i}}$ the $J+1$-dimensional basis vector for the observed treatment $D_{i}$. 

\item Compute $\hat{\pi}_{CAIPWL}=\text{argmax}_{\pi \in \Pi} \hat{Q}_{CAIPWL}(\pi)$. 
\end{enumerate}
The classes of policy over which maximizing the value-function can be numerous. In their work, the authors consider a decision-tree policy class providing an application to real data.   

\section{Online optimal policy learning}
Unlike offline policy learning, which involves learning from a fixed dataset, online policy learning takes place in an ongoing, interactive manner. In this approach, an agent or social planner continuously learns and updates the optimal policy to undertake by interacting with the environment. In online learning, one trains the model incrementally by feeding it data instances sequentially, either individually or by small groups called \textit{mini-batches} (Géron, 2022). 

The core idea of online policy learning is to consistently update the optimal policy whenever a new instance arrives, that is, when a new observation triplet becomes available. For instance, to estimate conditional means at each action/decision, one can employ \textit{online least squares} that are based on the gradient descent algorithm, which updates regression coefficients observation-by-observation in a sequential mode.

While offline policy learning can theoretically adopt a similar procedure, it necessitates re-estimating the optimal policy by refitting the model over the entire dataset, including the new incoming instance (this is called \textit{batch learning}). Typically, in offline learning, updates occur after a certain number of new instances are available. Consequently, for a certain sequential span, offline learning can use the same predicting mapping across several new instances until a decision is made to retrain the model (\textit{sequential batch learning}).
Nevertheless, in a non-Big Data setting, it is possible to refit the model observation-wise, providing a continuous update of the optimal policy. 

\begin{figure}[h!]
\centering
\includegraphics[width=15cm]{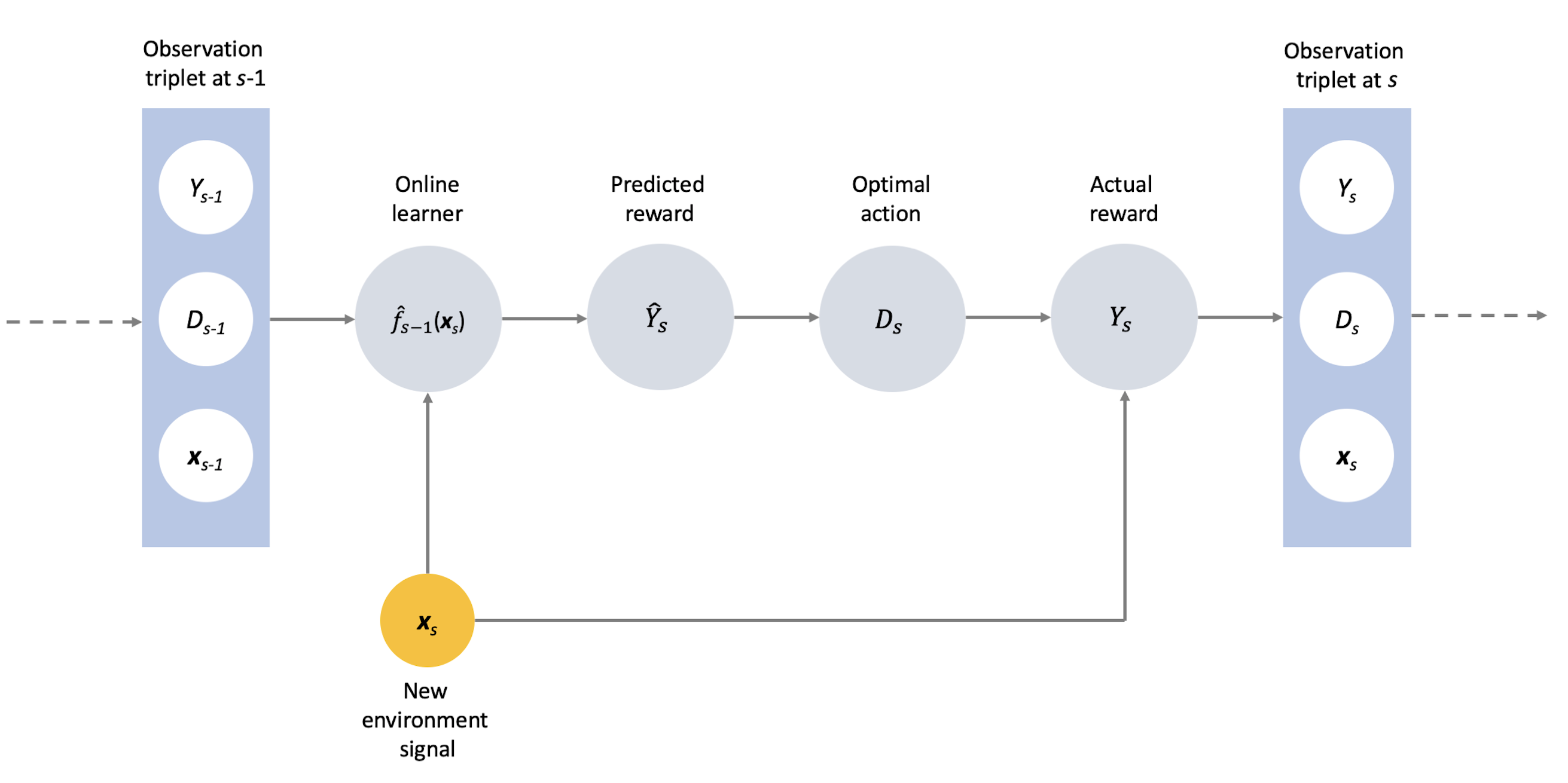}
\caption{A heuristic representation of the model's architecture.}
\label{fig:fig0}
\end{figure}

For the sake of clarity, It seems useful to present a heuristic representation of the type of online learning architecture applied to our context, where we consider the first-best optimal policy solution as reference. This example refers to an agent or a social planner taking decisions on the basis on an environmental signal. Therefore it encompasses both modes of OPL application, as  pointed out in the introduction. Figure \ref{fig:fig0} shows such architecture by clearly setting out the reinforcement learning nature of our model. Let's comment on this architecture.
We consider an agent or social-planner embedded in a given environment who has a specific task to carry out. In this setting, at decision round $s$, and for a certain configuration $\textbf{x}_{s}$ of the environment, the agent or the social planner has to come up with a new action/decision $D_{s}$ out of a finite set of actions/decisions on the basis of the generated expected reward $\hat{Y}_{s}$.   

Inherited from the past -- that is, at action/decision round $s-1$ -- the learning process entailed by this architecture starts by considering the availability of an observation triplet $\{Y_{s-1},D_{s-1},\textbf{x}_{s-1}\}$. In this triplet, $Y_{s-1}$ is the actual reward at $s-1$, $D_{s-1}$ is the action/decision undertaken at $s-1$, and $\textbf{x}_{s-1}$ is the vector of environment signals at $s-1$. Through an online learning process, a machine can train the model over this triplet using a specific learner (for example, a random forest algorithm) thus obtaining the predictor $\hat{f}_{s-1}(\cdot)$ which produces a predicting mapping between the environment signal $\textbf{x}$ and the expected reward $\hat{Y}$ for each selected action/decision $D$. At the new action/decision round $s$, a new environment signal $\textbf{x}_{s}$ shows up, and the model can estimate -- for each action/decision -- the predicted reward $Y_{s}$ at round $s$ using the pre-estimated mapping $\hat{f}_{s-1}(\cdot)$. 

In line with what we have seen for the offline learning, the model \textit{first-best} policy solution selects the \textit{best} action $D_{s}$ to undertake as the one predicting the largest expected reward. After undertaking this action, the actual reward $Y_{s}$ is returned, thus allowing for the availability of a new observed triplet $\{Y_{s},D_{s},\textbf{x}_{s}\}$ for the action/decision round $s$. The learning process continues to take place starting this time from the new triplet and finally providing a third triplet at action/decision round $s+1$, and so forth.               

\subsection{Estimation of the first-best policy}

Similarly to the offline learning, a simple procedure can be set out to estimate the first-best policy. Assume that assumptions A1 and A2 hold and suppose to have the following i.i.d. sample of observations $\{Y_{s}, D_{s}, \textbf{x}_{s}\}$, with $s=1, \dots,S$,  and $D_{s} = 0, 1, \dots,J$, then an estimation of $\mu_{s}(j,\textbf{x}_{s})$ can be obtained using a prediction of $Y_{s}$ obtained from a machine learning regression of $Y_{s}$ on $\textbf{x}_{s}$ in the subgroup of observations having $D_{s}=j$. In this way, we have a consistent estimate of all the counterfactuals for each observation round $s$.  

Suppose now to have a new observation $\textbf{x}_{i,s+1}$ and, given it, we would like the agent to select a specific action to undertake. This can be carried out based on procedure 2 (see below). 

\begin{table} [h!]
\centering
\begin{tabular}{>{\hspace{0pt}}m{0.94\linewidth}}  
\rule{\linewidth}{0.4pt} 
\textbf{Procedure 2.} \textit{Optimal action selection under assumptions A1 and A2}                                                                                                                                                                

\begin{itemize}
\item Generate the mapping between $Y_{s}$ and $\textbf{x}_{s}$ for each $D_{s} = 0, 1, \dots,J$ using a specific learner, and obtain the following set of $J$ predictors: 
$$\mathcal{M}_{s} = \{\hat{\mu}_{s}(0,\textbf{x}_{s}), \{\hat{\mu}_{s}(1,\textbf{x}_{s}), \dots , \hat{\mu}_{s}(j,\textbf{x}_{s}), \dots , \hat{\mu}_{s}(J,\textbf{x}_{s})\}$$

\item Given a new environment signal $\textbf{x}_{s-1}$, evaluates the previous set of predictions at $s+1$, thus getting:
$$\mathcal{M}_{i,s+1} = \{\hat{\mu}_{i,s+1}(0,\textbf{x}_{i,s+1}), \{\hat{\mu}_{i,s+1}(1,\textbf{x}_{i,s+1}), \dots , \hat{\mu}_{i,s+1}(j,\textbf{x}_{i,s+1}), \dots , \hat{\mu}_{i,s+1}(J,\textbf{x}_{i,s+1})\}$$

\item Select the best action to undertake at $s+1$ according to this rule:
$$ j_{s+1}^{*} =  \{j: \text{max}\{\mathcal{M}_{i,s+1}\}, j = 1, 0, \dots , J\}  $$  
\end{itemize}
\rule{\linewidth}{0.4pt} 
\end{tabular}
\end{table}

\begin{figure}[t]
\centering
\includegraphics[width=15cm]{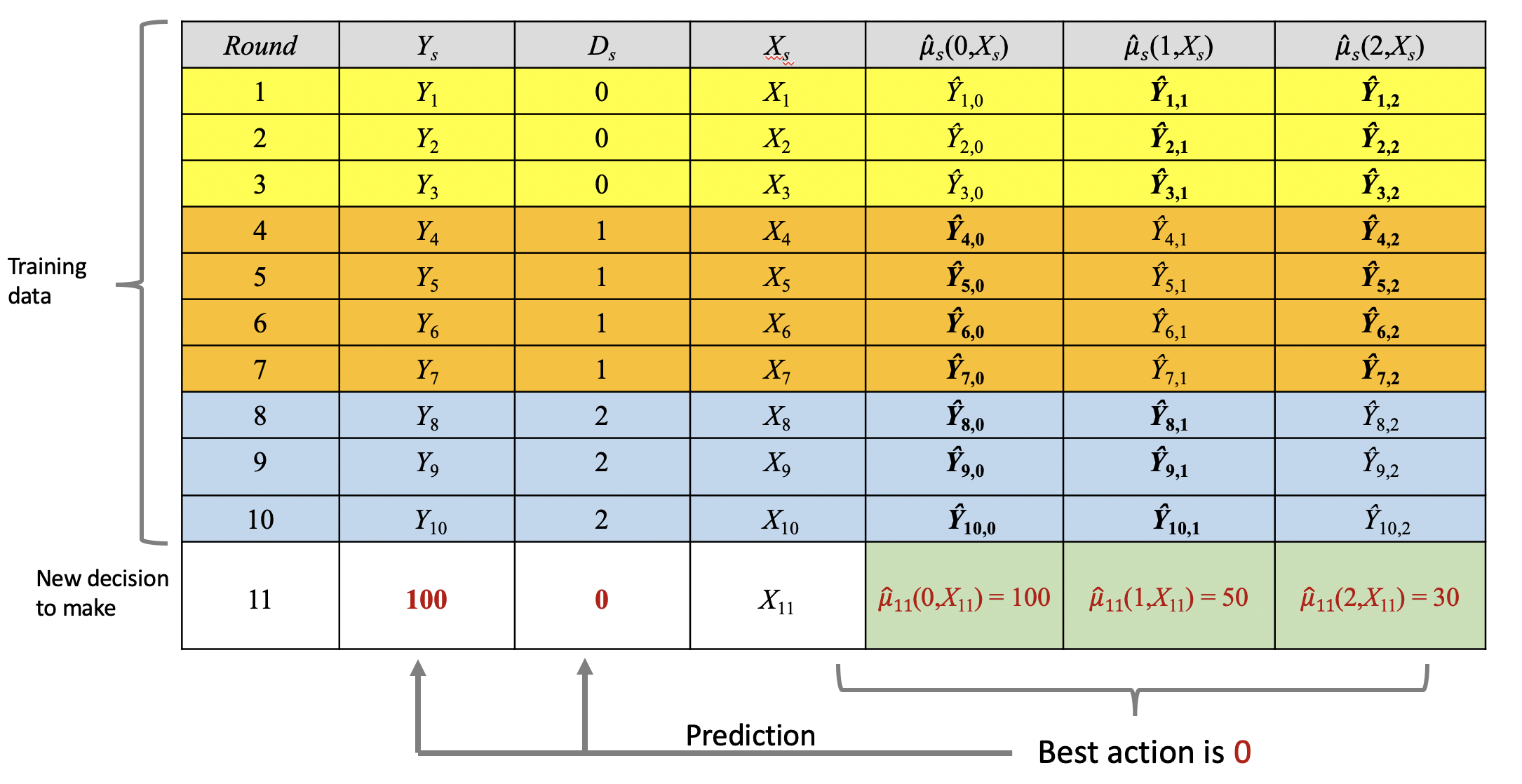}
\caption{Computation of the optimal choice when a new environment signal comes up according to Procedure 2, under assumptions A1 and A2.}
\label{fig:fig1}
\end{figure}

Figure \ref{fig:fig1} shows an example of the application of Procedure 2 when a new instance from the environment at round $s=11$ comes up, and when only three actions/decisions are available, either  action 0, action 1, or action 2. For this new instance, the signal from the environment is $X_{s+1}=X_{11}$, and the best choice to select is ``0'' as it entails the largest expected value of the reward (equal to 100). Observe that, as a consequence of assumption A1, $\hat{\mu}_{s}(0,X_{11})$ is the prediction at $X_{s+1}=X_{11}$ obtained from regressing the vector $\{ Y_1,Y_2, Y_3 \}$ on the vector $\{ X_1,X_2, X_3 \}$;  $\hat{\mu}_{s}(1,X_{11})$ is the prediction at $X_{s+1}=X_{11}$ obtained from regressing the vector $\{ Y_4,Y_5, Y_6, Y_7 \}$ on the vector $\{ X_4,X_5, X_6, X_7 \}$; finally, $\hat{\mu}_{s}(2,X_{11})$ is the prediction at $X_{s+1}=X_{11}$ obtained from regressing the vector $\{ Y_8,Y_9, Y_{10} \}$ on the vector $\{ X_8,X_9, X_{10} \}$. 

This procedure optimally selects the best action based on the expected reward. However, the expected reward cannot be a credible reference for optimal choice selection when the reward distribution is highly spread. This has to do with the presence of reward uncertainty, an aspect deserving special attention as it can remarkably affect the ultimate choice to select (Manski, 2013).

\section{Optimal decision under reward uncertainty}
In an uncertain environment, the returns from undertaking specific actions are associated to risk and uncertainty. In such a context, choosing, let's say, action A instead of action B depends not only on the average return of each option, but also on the uncertainty in getting such return. Therefore, decision-making must ponder the return and its related variability.   

\begin{figure}[t]
\centering
\includegraphics[width=15cm]{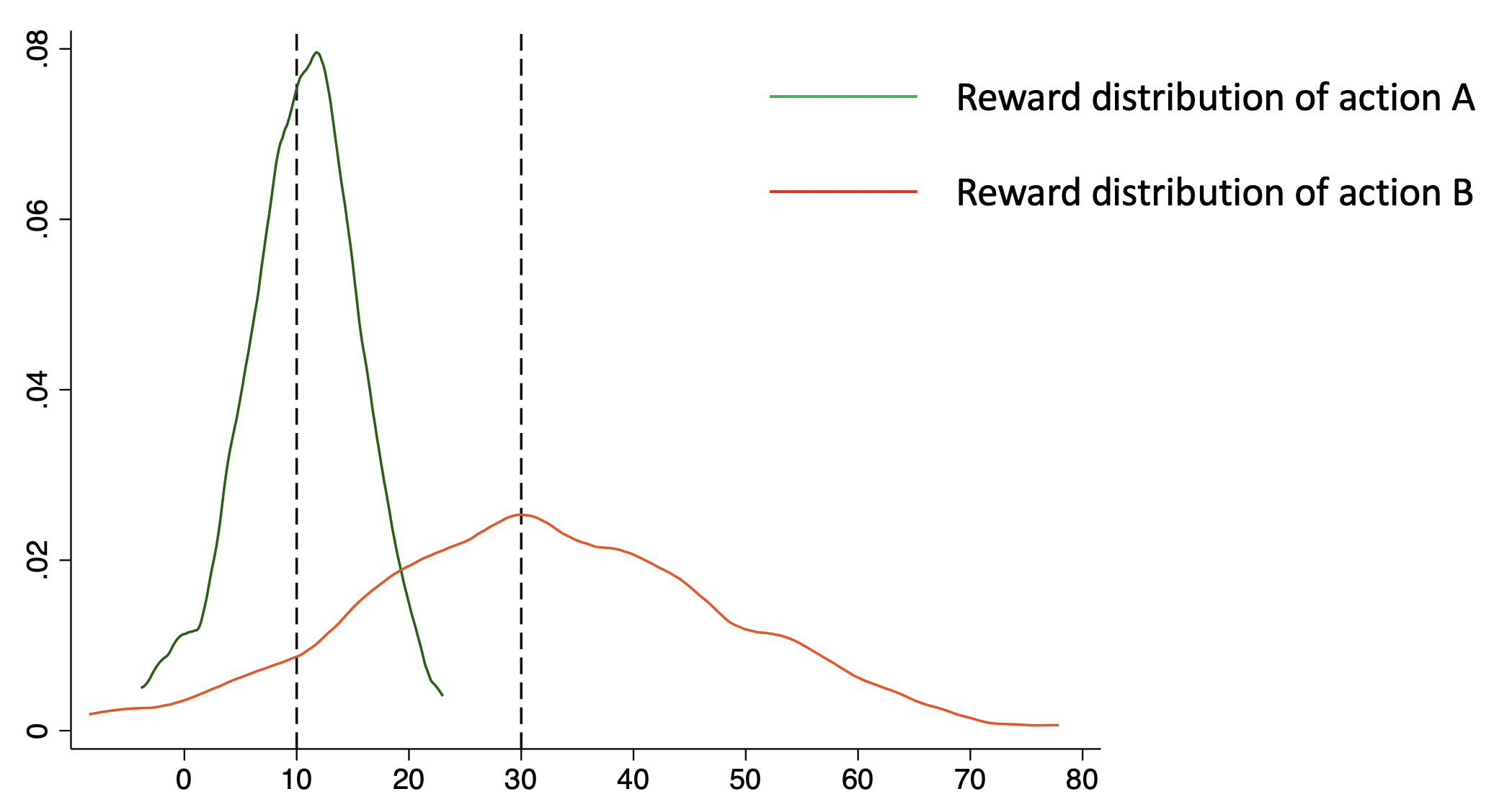}
\caption{Reward distribution and uncertainty realtive to two action, A and B. Action A provides a lower average return, but with smaller uncertainty. Action B provides a higher average return, but with larger uncertainty.}
\label{fig:fig2}
\end{figure}

Figure \ref{fig:fig2} shows the reward distribution and related uncertainty for two actions, A and B. We see that action A provides a lower average return, but with smaller uncertainty, whereas action B provides a higher average return but with larger uncertainty. In this case, it is not clear what action should be optimally undertaken, as a trade-off between expected reward and uncertainty takes place.

The issue has been well-recognized by a recent stream of multi-armed bandit literature focusing on risk-adverse agents taking decisions not only on the basis of average reward, but also incorporating reward's uncertainty in their choice measured using, for example, the variance of the reward distribution (Sani et al., 2012). When the objective function incorporates risk, traditional  algorithms trading-off exploration and exploitation with the aim of minimizing the policy regret, can take a different form and can have different asymptotic performance compared to traditional risk-neutral algorithms. 

Sani et al. (2012) address what they call in their paper the \textit{mean-variance} multi-armed bandit problem (Markowitz, 1952). Working on an exploration/exploitation learning setup, the authors investigate the role of reward uncertainty \textit{arm-wise}, that is, by defining for each arm $j$ the following mean-variance objective function:
$$ MV_{j} = \sigma^{2}_{j} - \rho \mu_{j} $$
where $\sigma^{2}_{j}$ is the variance, $\mu_{j} $ the mean of the reward distribution $F(Y_{j})$, and $\rho$ is the coefficient of absolute risk tolerance. 

The best arm, $j^{*}$, is the one minimizing the mean-variance, that is:
$$j^{*} = \text{argmin}_{(0,1,\dots,J)} \{MV_{j}\}$$
We can notice that when $\rho \rightarrow \infty$, the mean-variance of arm $j$ leads to the standard expected reward maximization of traditional multi-armed bandit problems. When $\rho=0$, the mean-variance criterion becomes equivalent to minimizing the variance. In this latter case, the objective becomes variance minimization.

A recent paper by Cassel et al. (2023) generalizes the Sani et al. (2012) approach by investigating the interplay between arm reward distributions and risk-adjusted performance metrics which includes conditional value-at-risk, mean-variance trade-offs, Sharpe ratio, and other risk metrics.

The literature on multi-armed bandit with observational data, which is the one we refer to in this paper, has given less attention to the problem of estimating policy risk. Recently, however, three papers have contributed to this subject by focusing on the estimation of the reward uncertainty under different policy scenarios. 

Chandak et al. (2021) provide consistent estimation of the offline  variance of the return associated to the policy $\pi$ defined as:
\begin{equation} \label{eq:var1}
\sigma^{2}(\pi) = \text{Var}[Y(\pi(\textbf{x})]
\end{equation}
Indeed, the return distribution is not only characterized by a central measure like the average reward of equation (\ref{eq:valfun}), but also by variability around this central measure. 
\\
\\
\noindent 
\rule{\linewidth}{0.4pt} 
\textbf{Example 2.} \textit{OPL with risk-adjusted linear reward and threshold-based policy class}. 
\\
Consider the same setting of example 1. In this case, we saw that:
\begin{equation} \label{eq:ex2}
Y = \alpha(c) \cdot \pi(X) + \epsilon
\end{equation}
where $\epsilon$ is pure random variable uncorrelated with $X$, with zero mean and finite variance. 
As policy class, we considered the \textit{threshold-based} policy rule:
\begin{equation} \label{eq:treshPolicy}
\pi(X) = 1[X<c]
\end{equation}
where $c$ is a constant. We proved that the average reward is:
\begin{equation} \label{eq:ex4}
\text{E}(Y) = \alpha(c) \cdot \text{E}(1[X<c])= \alpha(c) \cdot \text{Prob}(X<c)=\alpha(c) \cdot F_{X}(c)
\end{equation}
where $F_{X}(c)$ is the c.d.f. of $X$ evaluated at $c$. 
Now, we can estimate also the variance of $Y$ as:
\begin{equation} \label{eq:ex4}
\text{Var}(Y) = \alpha(c)^{2} \cdot \text{Var}(1[X<c]) + \sigma^{2}_{\epsilon}=
\alpha(c)^{2} \cdot F_{X}(c)[1-F_{X}(c)]+\sigma^{2}_{\epsilon}
\end{equation}
We can thus define a risk-adjusted expected reward as:
\begin{equation} \label{eq:ex5}
\gamma(c) = \frac{\text{E}(Y)}{\text{Var}(Y)} = \frac{\alpha(c) \cdot F_{X}(c)}{\alpha(c)^{2} \cdot F_{X}(c)[1-F_{X}(c)]+\sigma^{2}_{\epsilon}}
\end{equation}
We define the optimal policy as:
\begin{equation} \label{eq:ex6}
\pi^{*}(X) = 1[X<c^{*}]
\end{equation}
where:
$$c^{*}=\text{argmax}_{c}[\gamma(c)]$$
\rule{\linewidth}{0.4pt} 
In OPL with observational data, scholars aim to estimate the overall variance of the policy. However, in this paper, we propose a pretty different approach closer to OPL with online learning. 
Indeed, instead of focusing on the estimation of the overall total variance of the policy, we focus our attention on the estimation of the \textit{conditional variance}, and introduce specific risk preferences. Let's delve into this approach. 

Conditional uncertainty can be measured via the conditional variance, which is the variance of the distribution of $Y|\textbf{x}$. The formula of the conditional variance is:
\begin{equation} \label{eq:variance}
\text{Var}(Y|\textbf{x}) = E[Y – \text{E}(Y)|\textbf{x}]^2 = \text{E}(Y^2|\textbf{x}) – {\text{E}(Y|\textbf{x})}^2 
\end{equation}
We proceed action-wise and step-by-step, as in online learning. Therefore, at round \textit{s}, we estimate the conditional variance associated to arm $j$ as:
$$\sigma^{2}_{s}(j,\textbf{x}_{s}) = \text{Var}(Y_{s}|D_{s}=j,\textbf{x}_{s})$$
which can be easily estimated as the difference between two conditional means as in formula (\ref{eq:variance}):
\begin{equation} \label{eq:variance2}
\hat{\sigma}^{2}_{s}(j,\textbf{x}_{s}) = \hat{\text{E}}(Y^{2}_{s}|D_{s}=j,\textbf{x}_{s}) - \hat{\text{E}}(Y_{s}|D_{s}=j,\textbf{x}_{s})^2
\end{equation}
where the conditional means in the RHS can be estimated using specific machine learning techniques. Thus, the optimal action to select at $s+1$ given the signal $\textbf{x}_{i,s+1}$ depends on the pair:
$$ [ \hat{\mu}_{i,s+1}(j,\textbf{x}_{i,s+1})  ,  \hat{\sigma}_{i,s+1}(j,\textbf{x}_{i,s+1})] $$
and on the preferences between return and risk. Observe that $\hat{\sigma}_{i,s+1}(\cdot)$ is the estimated standard deviation. 

We assume a \textit{risk-averse} decision-maker, i.e. one preferring lower levels of risk for a given level of return. A utility function for a risk-averse decision-maker would reflect this preference by assigning a lower utility value to actions with higher levels of risk.
Risk-averse preferences can be modeled through a utility function whose arguments are the conditional average reward and the conditional standard deviation. Here we consider two settings: (i) linear risk-averse preferences, and (ii) quadratic risk-averse preferences. Two actions can have different preferential ordering according to the specific type of preferences assumed.
\\
\\
\textit{Linear risk-averse preferences}. The utility function is equal to the ratio between the conditional average reward and the conditional standard deviation:  
\begin{equation} \label{eq:linear_pref}
U_{it,L}=\frac{\hat{\mu}_{s}}{\hat{\sigma}_{s}}
\end{equation}
implying, by equalizing $U_{it,L}$ to a constant $k$, a linear indifferent curve:
\begin{equation} \label{eq:linear_pref}
\hat{\mu}_{s} = \hat{\sigma}_{s} + k
\end{equation}
\\
\textit{Quadratic risk-averse preferences}. The utility function is equal to the ratio between the conditional average reward and the squared value of the conditional standard deviation:
\begin{equation} \label{eq:linear_pref}
U_{it,Q}=\frac{\hat{\mu}_{s}}{\hat{\sigma}_{s}^{2}}
\end{equation}
implying, by equalizing $U_{it,Q}$ to a constant $k$, a quadratic indifferent curve:
\begin{equation} \label{eq:linear_pref}
\hat{\mu}_{s} = \hat{\sigma}_{s}^{2} + k
\end{equation}

\begin{figure}[t]
\centering
\includegraphics[width=10cm]{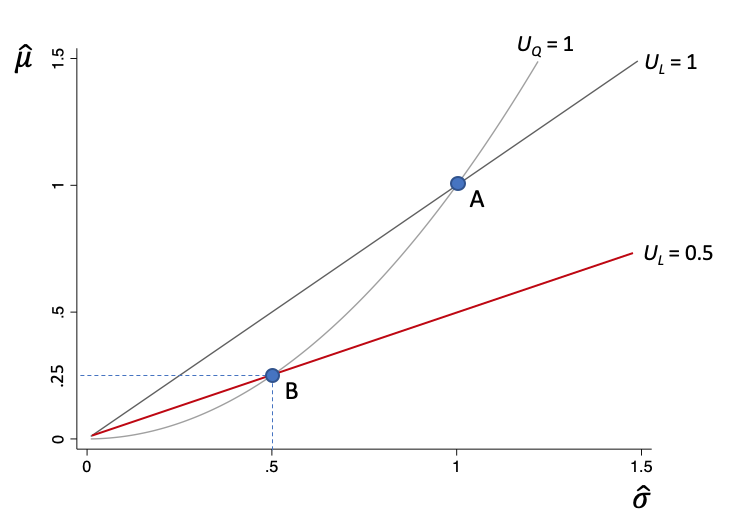}
\caption{Example of actions' preferential ordering. Under a linear risk-averse preferences, the agent prefers action A over action B. Under a quadratic risk-averse preferences, the agent is indifferent between action A and B.}
\label{fig:fig3}
\end{figure}

Figure \ref{fig:fig3} shows an example of actions' preferential ordering. We can easily see that according to linear risk-averse preferences, the agent turns out to prefer action A over action B. On the contrary, according to quadratic risk-averse preferences, the agent is indifferent between action A and B.

We can conclude that, when comparing alternative actions under different risk-averse preferences, the preferential ordering can change.\footnote{For example, an alternative that is preferred under a logarithmic utility function may not be preferred under a power utility function. This is because the power utility function assigns a higher weight to extreme outcomes, which means that the potential losses associated with the alternative may outweigh any potential gains.} It is thus intriguing to explore the extent to which different attitudes of policymakers towards risk can significantly influence the optimal actions chosen and the corresponding average regret. In the next section, we delve into this subject by considering a real policy context, employing the risk-adjusted framework described above and using the first-best rule as our reference (optimal) decision algorithm.

\section{Application: optimal allocation of a job training policy}
As an illustrative example, I utilize the well-known LaLonde (1986) dataset \texttt{jtrain2.dta}, which was employed by Dehejia and Wahba (1999) to assess various propensity-score matching methods in an ex-post policy evaluation.
In their investigation, the authors aimed to estimate the impact of participating in a job training program administered in 1976 (indicated by the binary variable \texttt{train}, taking the value 1 for treated individuals and 0 for untreated) on real earnings in 1978 (variable \texttt{re78}) for a group of individuals in the United States. The dataset comprises a total of 445 observations, with 185 individuals treated and 260 untreated.

In our study, we designate the number of months of training (variable \texttt{mostrn}) as the treatment variable $D$, ranging from 0 to 24 months. The median for treated individuals is 21 months. Consequently, I construct a 3-arm set of actions:

\begin{itemize}
\item  Action 1: no training, $D=0$, $N_{0}=260$;
\item  Action 2: training between 1 month and 21 month, $D=1$, $N_{1}=107$;
\item  Action 3: training lasting from 22 to 24 months, $D=2$, $N_{2}=78$;.
\end{itemize}
where $N_{0}+N_{1}+N_{2}=N=445$.

I consider that the potential results of the target variable \texttt{re78} (which is the reward) are not influenced by the treatment variable $D$ - as defined earlier - once we control for the variables $\textbf{x}$.

Following the specifications outlined by Dehejia and Wahba (1999), I consider the following features: \texttt{age} (age in years), \texttt{agesq} (age squared), \texttt{educ} (years of schooling), \texttt{black} (an indicator variable for Black individuals), \texttt{hisp} (an indicator variable for being Hispanic), \texttt{married} (an indicator variable for marital status), \texttt{nodegr} (an indicator variable for a high school diploma), \texttt{re74} (real earnings in 1974), \texttt{re74sq} (real earnings in 1974 squared), \texttt{re75} (real earnings in 1975), \texttt{unemp74} (an indicator variable for being unemployed in 1974), \texttt{unemp75} (an indicator variable for being unemployed in 1975), and \texttt{u74hisp} (an interaction term between \texttt{unemp74} and \texttt{hisp}).

I consider two applications, based respectively on offline and online learning. 

\subsection{Offline learning}
 In this context, I work in an offline learning setting, where I create two distinct datasets: a \textit{training} dataset to learn the optimal policy, and a \textit{new} dataset to predict the optimal treatment allocation based only on the features of each unit. 

In order to create meaningful graphical representations, I have selected only 50 units at random for the training dataset, and for the new (unlabeled) dataset, I have chosen 30 units randomly. These new individuals will be assigned to different training actions solely based on their features. I consider three different settings: (i) risk-neutral, (ii) linear risk-adverse, and (iii) quadratic risk-adverse. For the estimation of the value function (and thus of the regret), I consider the three estimators outlined above in this paper, that is: Regression-Adjustment (RA), Inverse Probability Weighting (IPW), and Double-robust (DR).

\begin{figure}[t]
\centering
\includegraphics[width=15cm]{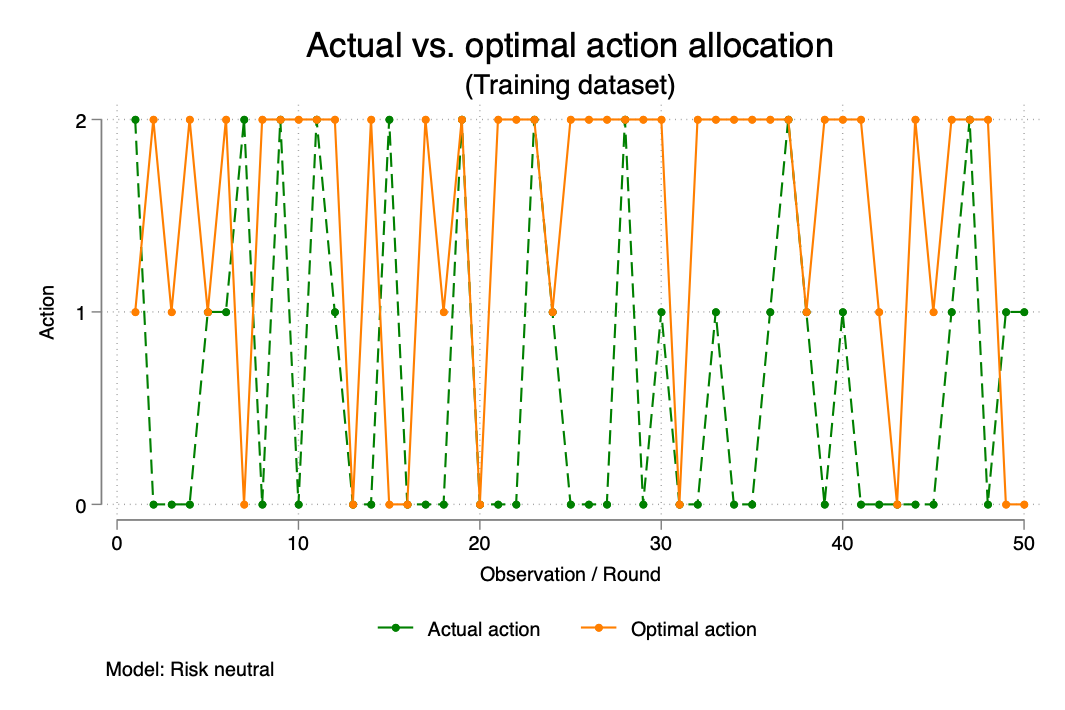}
\caption{Actual vs. optimal action allocation: risk-neutral setting. Offline learning.}
\label{fig:gr1}
\end{figure}

\vspace{0.7cm}
\noindent \textbf{Case 1.} \textit{Risk-neutral setting}.
We set out by applying the optimal action according to the algorithm listed in Procedure 2. Figure \ref{fig:gr1} plots the actual versus the optimal class allocation. By considering the matches -- i.e., cases in which the actual and the optimal individuals' allocation to the different classes coincide -- we can see that only the 30\% of the 50 individuals were allocated to the expected optimal class. All the remaining 70\% were allocated to the wrong class:   
\begin{verbatim}
    Variable |        Obs        Mean    Std. dev.       Min        Max
-------------+---------------------------------------------------------
      _match |         50          .3      .46291          0          1
\end{verbatim}

\begin{figure}[t]
\centering
\includegraphics[width=15cm]{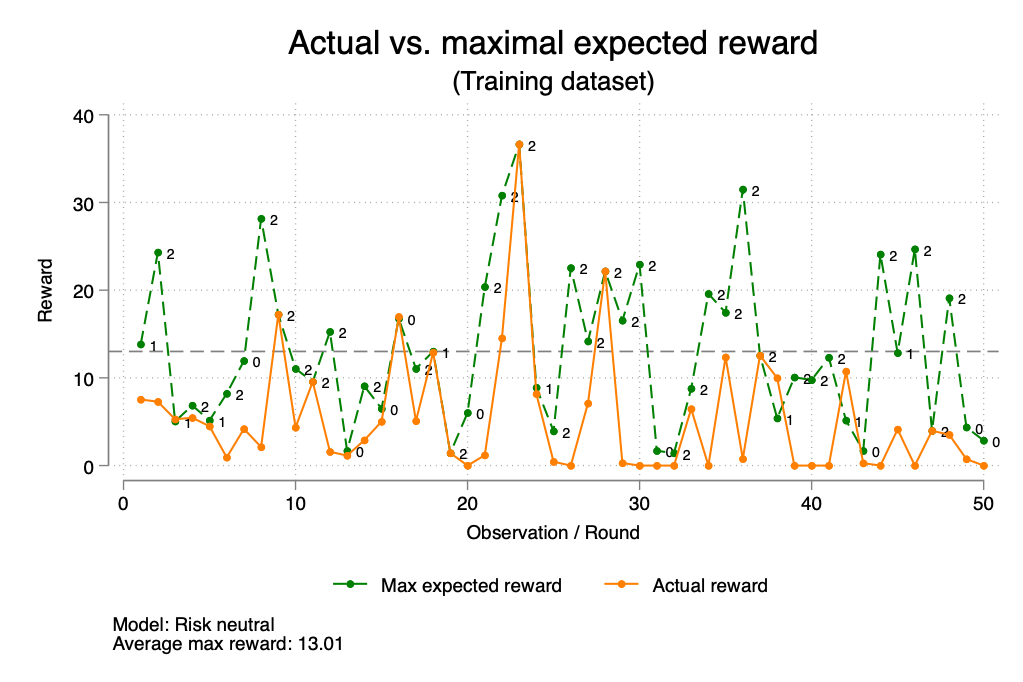}
\caption{Actual vs. optimal expected reward: risk-neutral setting. Offline learning. The number close to the point indicates the optimal class.}
\label{fig:gr2}
\end{figure}

When it comes to the estimation of the average regret, we have to contrast the actual with the maximum expected reward. For the training dataset, the two rewards are plotted in figure \ref{fig:gr2}, where the maximum expected reward dominates for pretty every individual the actual reward. For estimating the average regret of the policy, we contrast the estimation of the value function at the current policy with the value function estimated at the optimal policy using the RA, IPW, and DR estimators. We obtain that:       
\begin{verbatim}
--------------------------------
Regret RA = 8.891423
Regret IPW = 3.7557106
Regret DR = 7.3346037
--------------------------------
\end{verbatim}
We see that the regret is positive and quite large, going from 3.75 for the IPW estimator, to 7.33 for the DR. This can be interpreted as an average \textit{loss of welfare} due to the wrong allocation of individuals into classes of different training duration.    

\vspace{0.7cm}

\begin{figure}[t]
\centering
\includegraphics[width=15cm]{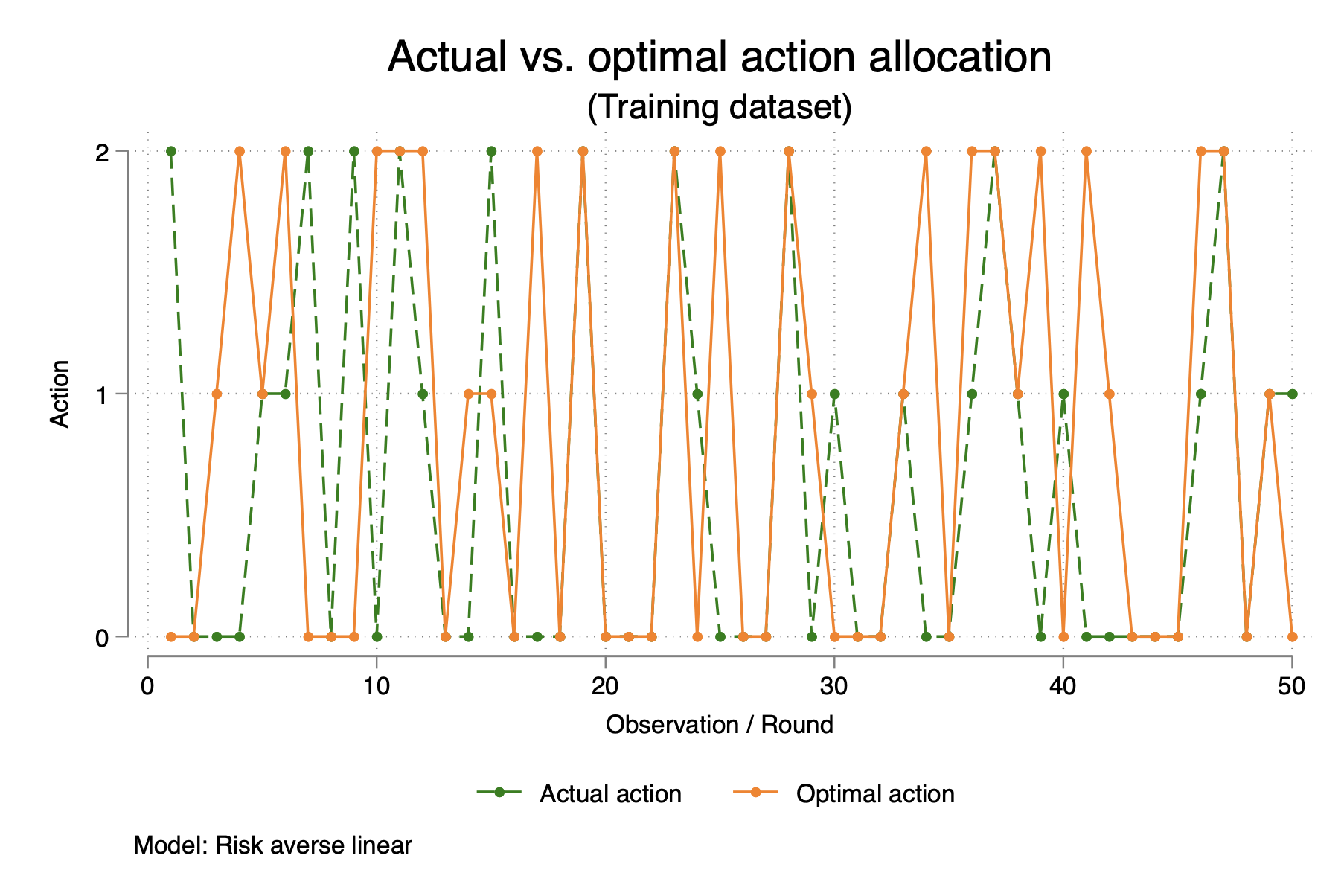}
\caption{Actual vs. optimal action allocation: risk-adverse linear setting.}
\label{fig:gr3}
\end{figure}

\noindent \textbf{Case 2.} \textit{Risk-adverse linear setting}.
Figure \ref{fig:gr3} plots the actual versus the optimal class allocation in the case of a policymaker with \textit{linear} risk-adverse preferences. In this setting, the share of matches grows up to 54\%, indicating a quite large increase in the right allocation of people to the different training classes:
\begin{verbatim}
    Variable |        Obs        Mean    Std. dev.       Min        Max
-------------+---------------------------------------------------------
      _match |         50         .54    .5034574          0          1
\end{verbatim}

\begin{figure}[t]
\centering
\includegraphics[width=15cm]{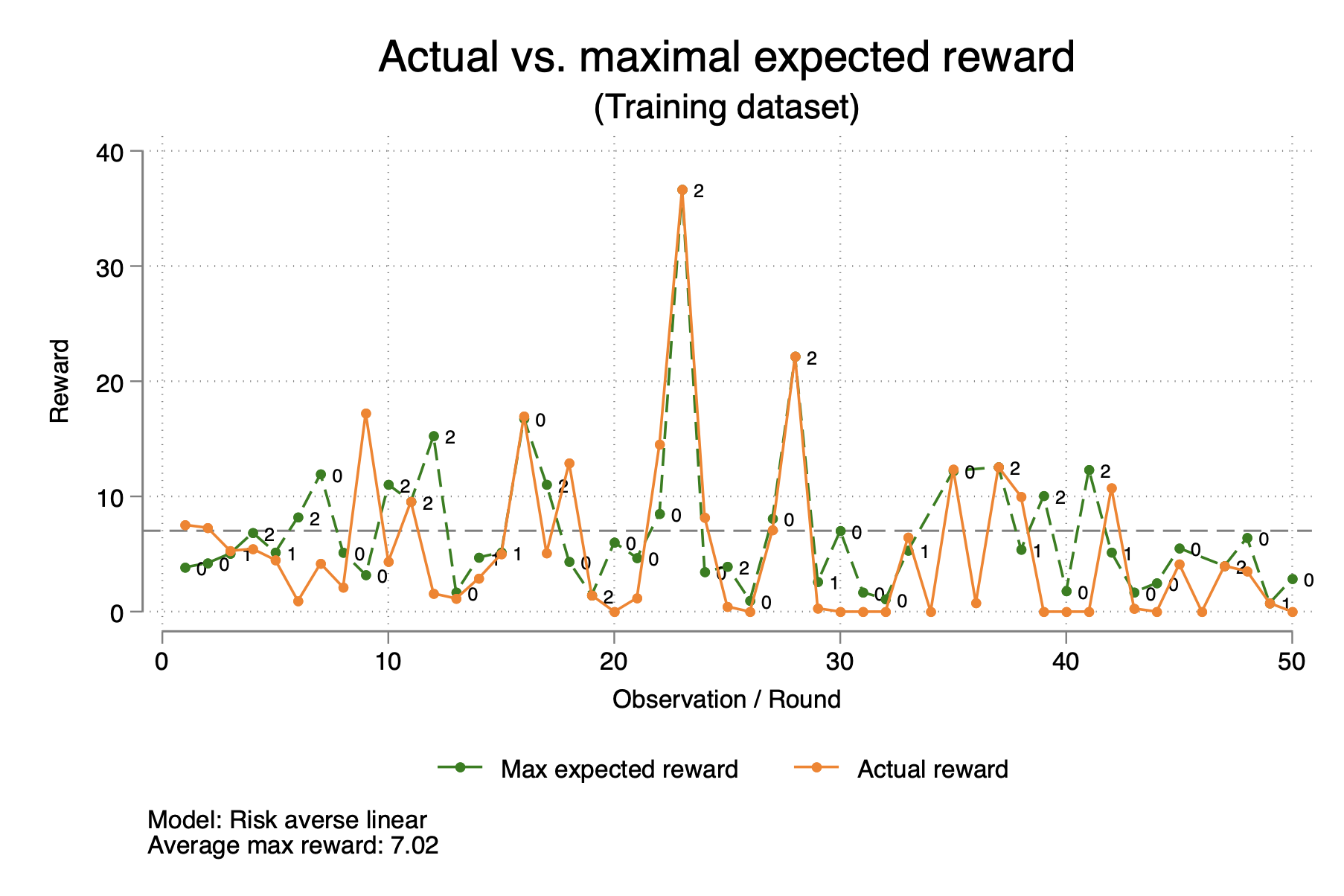}
\caption{Actual vs. optimal expected reward: linear risk-adverse setting. Offline learning. The number close to the point indicates the optimal class.}
\label{fig:gr4}
\end{figure}

We can also compute the average regret, which is equal to 3.41 for the RA, 0.55 for the IPW, and 2.58 for the DR:
\begin{verbatim}
--------------------------------
Regret RA = 3.4163201
Regret IPW = .55887842
Regret DR = 2.5841078
--------------------------------
\end{verbatim}
Finally, figure  \ref{fig:gr4} shows the actual versus the maximal expected reward in linear risk-adverse setting. Also in this case, we see that the optimal expected reward dominates pretty always the actual reward, thus confirming the finding set out in the previous table. 

\vspace{0.7cm}

\begin{figure}[t]
\centering
\includegraphics[width=15cm]{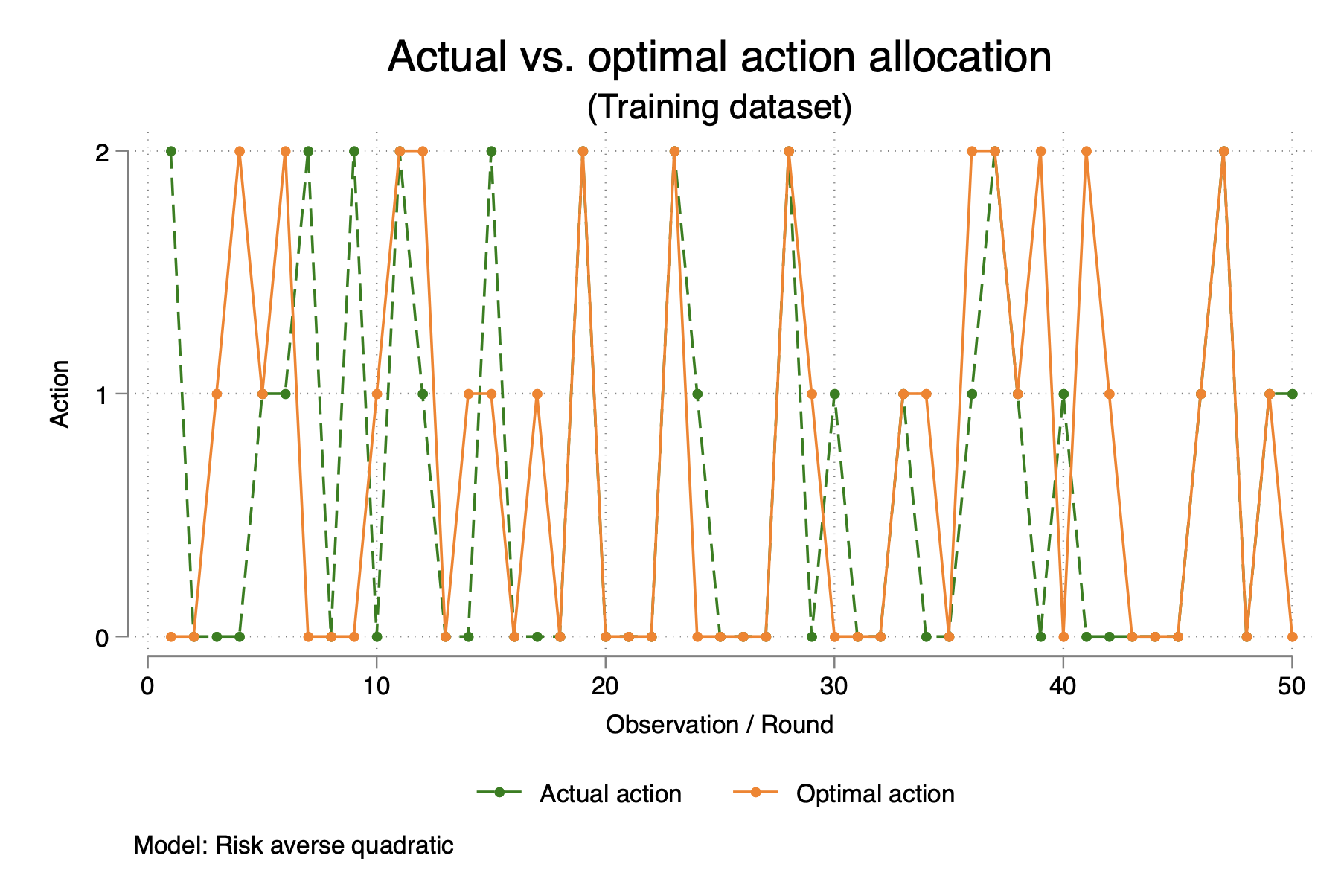}
\caption{Actual vs. optimal action allocation: quadratic risk-adverse setting. Offline setting.}
\label{fig:gr5}
\end{figure}

\noindent \textbf{Case 3.} \textit{Risk-adverse quadratic setting}.
Figure \ref{fig:gr5} plots the actual versus the optimal class allocation in the case of a policymaker with \textit{quadratic} risk-adverse preferences. In this setting, the share of matches is 58\%, indicating a quite large right allocation of people to the different training classes:
\begin{verbatim}
    Variable |        Obs        Mean    Std. dev.       Min        Max
-------------+---------------------------------------------------------
      _match |         50         .58    .4985694          0          1
\end{verbatim}

\begin{figure}[t]
\centering
\includegraphics[width=15cm]{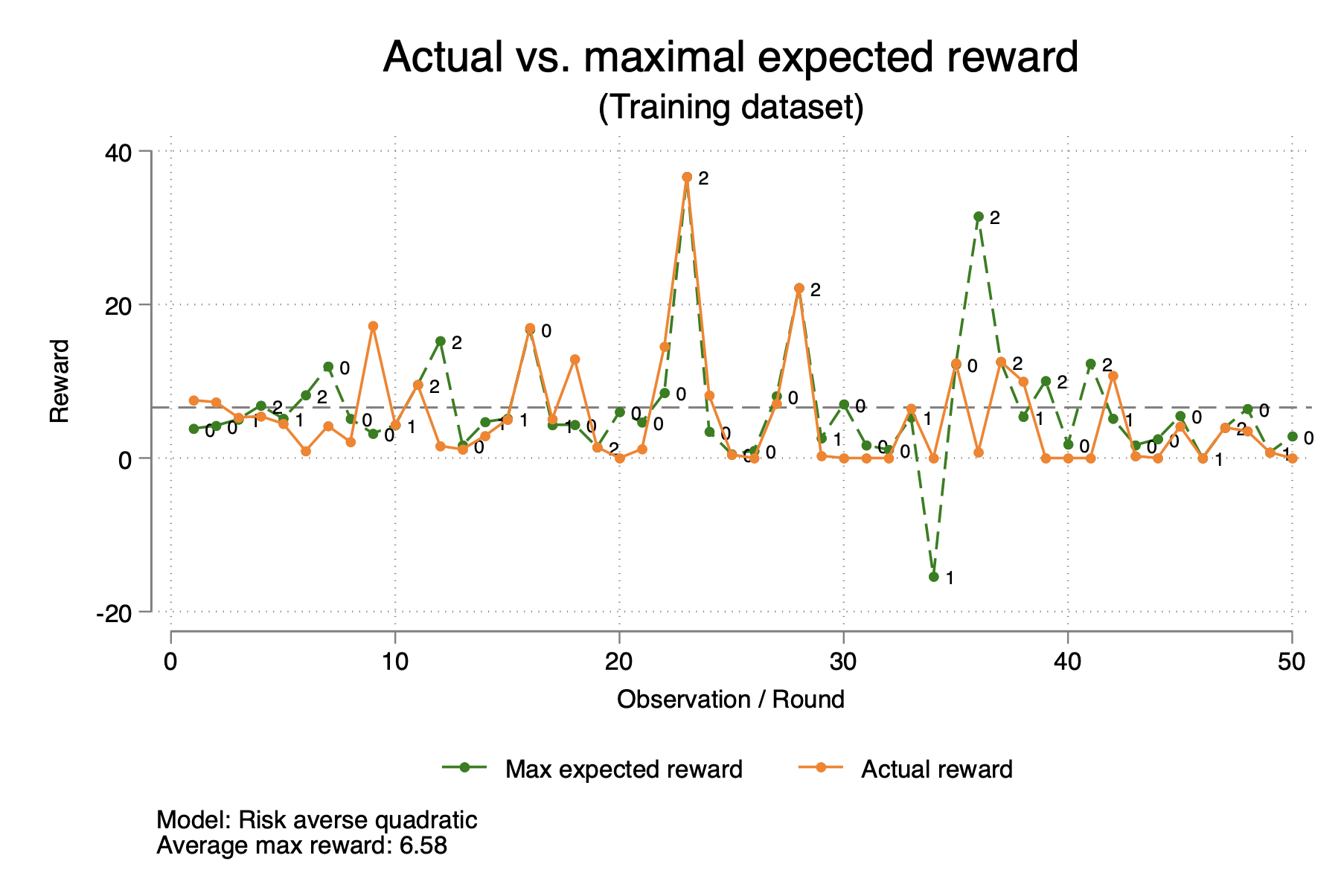}
\caption{Actual vs. optimal expected reward: quadratic risk-adverse setting. Offline learning. The number close to the point indicates the optimal class.}
\label{fig:gr6}
\end{figure}

We can also compute the average regret, which is equal to 0.03 for the IPW, 1.04 for the DR, and even negative (-5.08) for the RA, probably due to a large bias for this estimator:
\begin{verbatim}
--------------------------------
Regret RA = -5.0857218
Regret IPW = .03672314
Regret DR = 1.0449446
--------------------------------
\end{verbatim}
Figure  \ref{fig:gr6} shows the actual versus the maximal expected reward in the quadratic risk-adverse setting. In this case, the optimal expected reward still dominates the actual reward, thus confirming the finding set out in the previous table. 

\begin{figure}[t]
\centering
\includegraphics[width=15cm]{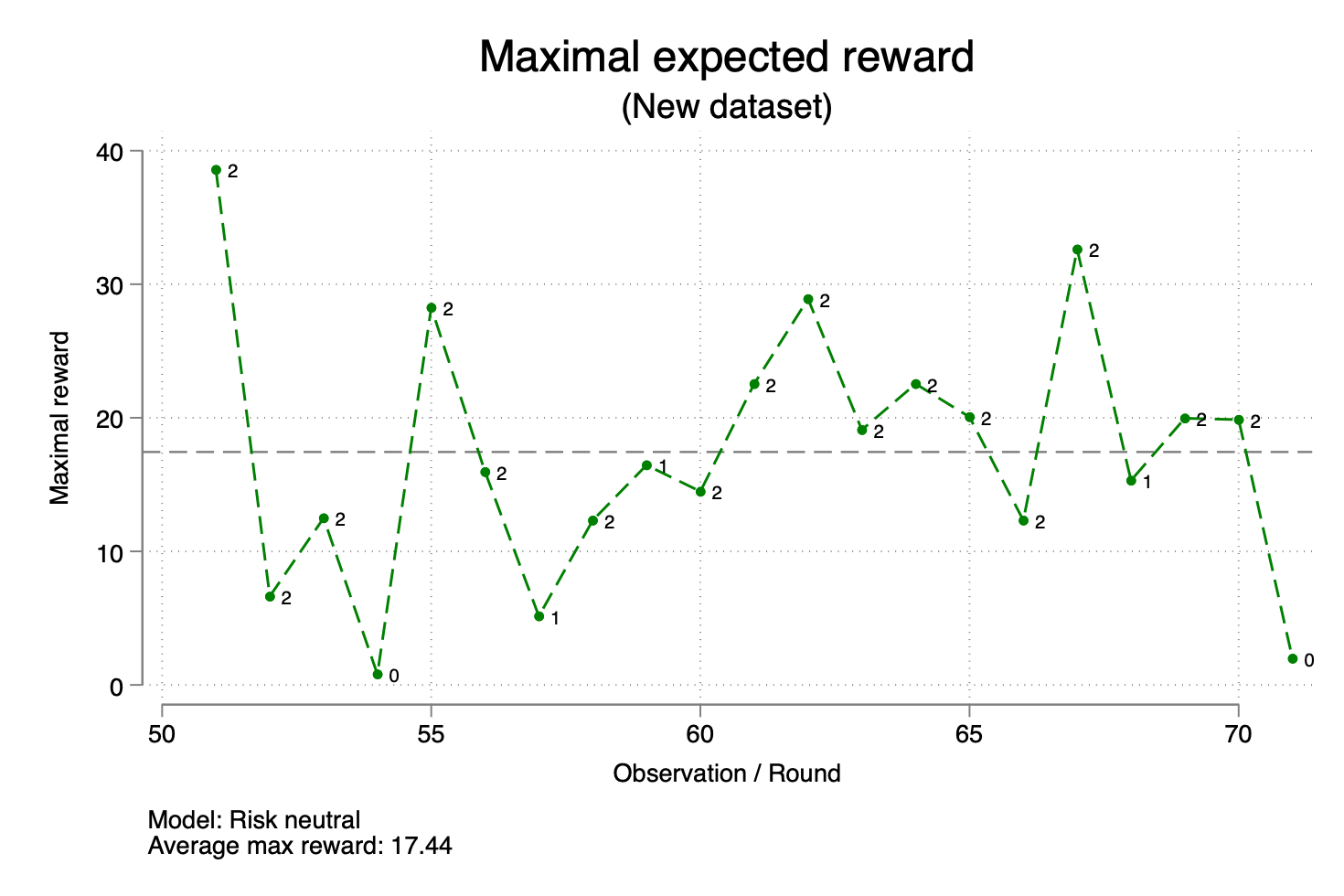}
\caption{Predicted optimal expected reward on new instances: risk-neutral setting. Offline setting.}
\label{fig:gr7}
\end{figure}

As a final step, it may be interesting to look at the predicted optimal class and expected reward on the new instances. For the sake of brevity, I consider only the risk-neutral setting. Figure \ref{fig:gr7} sets out the result.

\subsection{Online learning}
\begin{figure}[t]
\centering
\includegraphics[width=15cm]{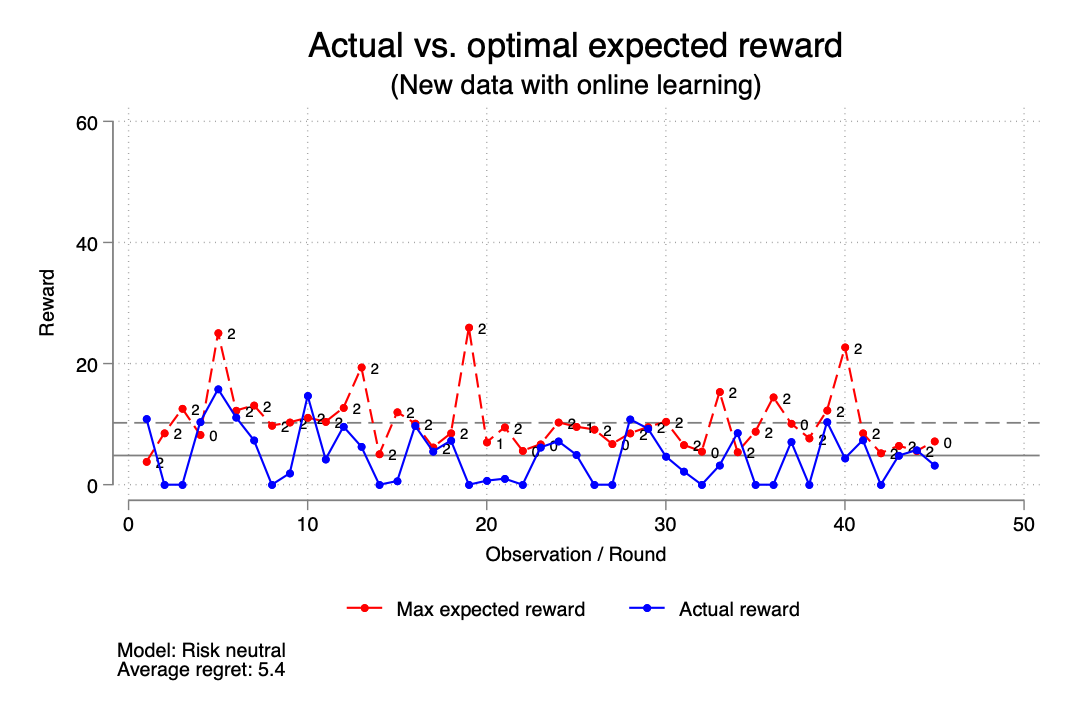}
\caption{Predicted optimal expected reward on new instances: risk-neutral setting. Online learning.}
\label{fig:gr8}
\end{figure}

In this section, utilizing the same dataset exploited in the previous section, I employ an online algorithm. In this scenario, the training dataset comprises 400 observations, with the remaining 45 serving as new instances. For conciseness, I assume a risk-neutral decision maker, and compute the regret using the RA estimator. Figure \ref{fig:gr8} illustrates the primary outcome by plotting the predicted optimal expected reward for the new instances. In contrast to the offline setting described earlier, the online approach retrains the model as long as a new an instance gets in, ensuring a continuous update of the regression coefficients. Consequently, this approach is computationally more expensive than offline learning, which necessitates only a single fit. But it is more precise. 

In the new dataset, the percentage of right treatment allocation is rather low, around 20\%:
\begin{verbatim}
    Variable |        Obs        Mean    Std. dev.       Min        Max
-------------+---------------------------------------------------------
     _ match |         45          .2    .4045199          0          1
\end{verbatim}
This confirms a rather large misallocations of units within the different treatment classes.
Finally, according to the RA estimation, the average estimated regret is equal to 5.4.   

\section{OPL potential failures}
As a data-driven decision making approach, OPL can incur fundamental limitations in its application. These limitations have to do with the invalidation of the two fundamental assumptions set at the basis of this approach, i.e. unconfoundedness, and overlapping. In what follows, I discuss the two situations separately.

\subsection{Problems of weak overlapping}

\begin{figure}[h!]
\centering
\includegraphics[width=15cm]{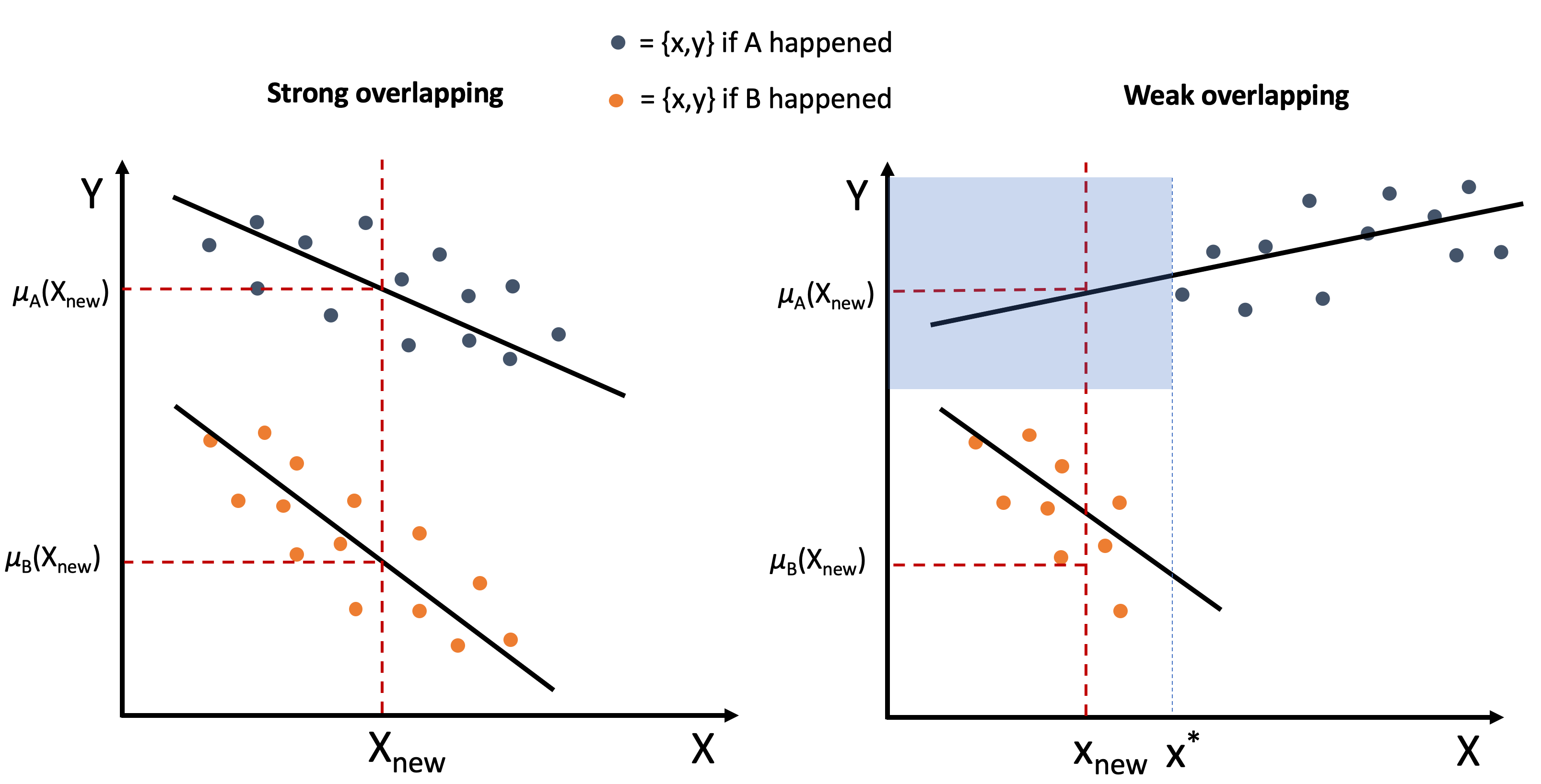}
\caption{Example of valid imputation (left-hand chart) and spurious imputation (right-hand chart) of $\mu_{A}(X_{new})$. Spurious imputation takes place when $X_{new} < X^{*}$ because of data sparseness due to weak overlap. Similarly, we can observe a spurious imputation of $\mu_{B}(X_{new})$ when $X_{new} > X^{*}$ due, again, to weak overlap.}
\label{fig:fig4}
\end{figure}

Figure \ref{fig:fig4} shows an example of a valid imputation of $\mu_{A}(X_{new})$ due to a good overlap (left-hand chart), and an example of spurious imputation of $\mu_{A}(X_{new})$ when $X_{new} < X^{*}$ because of data sparseness due to weak overlap (right-hand chart). In this latter case, the linear projection of the blue points is made in an area where only orange points are present. Therefore, this entails a spurious identification of $\mu_{A}(X_{new})$.     
 
\begin{figure}[t]
\centering
\includegraphics[width=15cm]{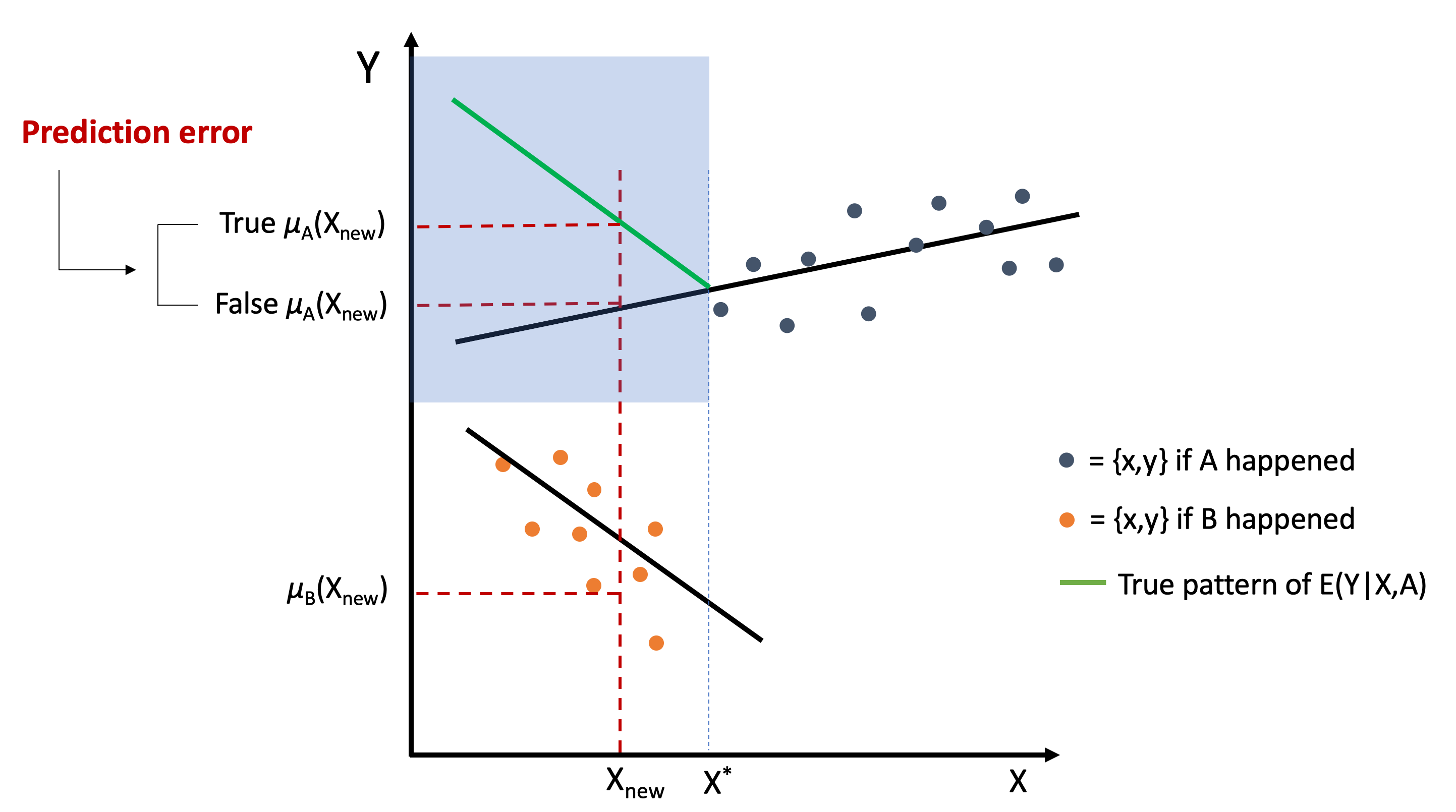}
\caption{The problem of weak overlap. When there is a weak overlap (i.e., sparseness) issue, the counterfactual cannot be correctly identified by data. In this case, we can make severe errors in predicting $\mu_{A}(X_{new})$. The green line is the true conditional expectation to estimate, but in the presence of weak overlap, we erroneously rely on the linear projection of the blue points.}
\label{fig:fig5}
\end{figure}

More clearly, figure \ref{fig:fig5}  shows the prediction error regarding the imputation of the conditional expectation $\mu_{A}(X_{new})$ that we can make in the presence of weak overlap. Indeed, while the green line represents the ``true'' conditional expectation we would like to impute, one erroneously commits an imputation error by relying on the linear projection of the blue points.    

\begin{figure}[h!]
\centering
\includegraphics[width=15cm]{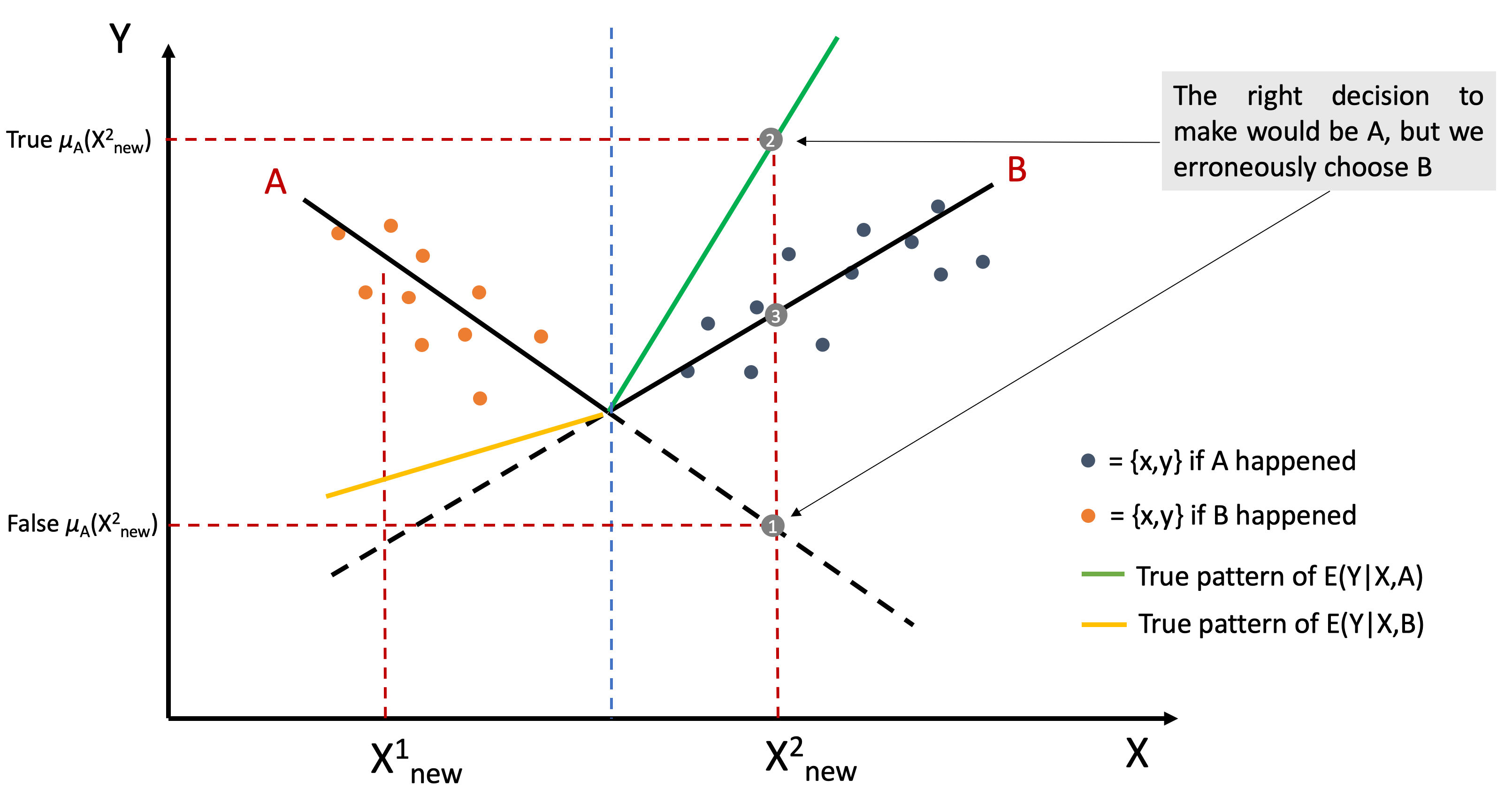}
\caption{Example of two-action inverted preference ordering due to the absence of overlap: when action A is selected, the true conditional mean is the green line, and the correct prediction at $X=X^{2}_{new}$ is in the gray point 2. Because of weak overlap (i.e., sparseness), the actual prediction at the value $X=X^{2}_{new}$ is in the gray point 1 which is however wrong. More importantly, the wrong prediction leads to invert the preferences, as action B is preferred to action A under wrong prediction (no overlapping), while A is preferred to B under correct prediction (overlapping).}
\label{fig:fig6}
\end{figure}

More critically, figure \ref{fig:fig6} shows an illustrative example of an inverted preference ordering between two actions due to weak overlap. In this case, we see that, when we select action A, the true conditional mean is the green line, and the correct prediction at $X=X^{2}_{new}$ is in the gray point 2. Because of weak overlap (i.e., sparseness), the actual prediction at the value $X=X^{2}_{new}$ is in the gray point 1 which is however wrong. More importantly, such wrong prediction leads to invert the preferences, as action B is preferred to action A under no overlapping, while A is preferred to B under overlapping.

The consequences of a data weak overlap can be severe, but in general it is never a problem of presence versus absence of overlap, but rather a problem of \textit{degree} of overlap. Fortunately, the degree of data overlap can be measured and tested, thereby obtaining some reliability measure regarding the quality of our imputations of the conditional means used for drawing the best decision (Busso, DiNardo, and McCrary, 2014).     

\subsection{Problems of weak unconfoundedness}

The unconfoundedness assumption (A1) assumes that, conditional on the knowledge of the environment (i.e., the vector $\textbf{x}_{s}$), there is statistical independence between the potential outcome when decision $j$ is selected and the decision $j$'s dummy. This entails conditional randomization of the undertaken choice, once the signal from the environment has been tapped. 

This assumption rules out the possible existence of other environmental components, $\tilde{\textbf{z}}_{s}$, having an effect on $Y_{s}(j)$ and simultaneously on $d_{s}(j)$ (\textit{confounders}). If such extra components exist, but are not observable in the data, we can no longer invoke decision's conditional randomization. This entails that the prediction of the optimal action could be highly affected by such \textit{hidden} confounders, thus making the conclusions about what is the best action to undertake potentially misled. Under weak unconfoundedness equation (\ref{eq:cmi1}) no longer holds, thereby having:

\begin{equation} \label{eq:NOcmi}
\mu_{s}(j,\textbf{x}_{s},\tilde{\textbf{z}}_{s}) \neq  \text{E}(Y_{s}| D_{s}=j ,\textbf{x}_{s}) 
\end{equation}
which implies that the counterfactual no longer can be estimated via the available data. Indeed, without unconfoundedness:

\begin{equation}
\text{E}(Y_{s}(j)|D_{s}=j,\textbf{x}_{s}) \neq
\text{E}(Y_{s}(j)|\textbf{x}_{s})  
\end{equation}
as the potential outcomes are now dependent of the decision dummy even if we condition over $\textbf{x}_{s}$. Therefore, relying on an estimation of the mapping identified by $\text{E}(Y_{s}| D_{s}=j ,\textbf{x}_{s})$ using whatever available learner would provide inconsistent  estimates of                                   
$\text{E}(Y_{s}(j)|\textbf{x}_{s}) $. 

Possible solutions to weak unconfoundedness can be:

\begin{itemize}
\item \textit{Collecting more data on the environment}. One way to address weak unconfoundedness is to collect more data on potential confounders. This may involve collecting additional contextual variables that are related to both the action selection and the reward.

\item \textit{Using methods robust to unobservable selection}. There are alternative methods to standard methods of causal inference that may be more robust to violations of weak unconfoundedness. For example, instrumental-variables (IV) analysis, or difference-in-differences (DID) analysis, are valid alternatives. IV estimation, however, requires the availability of an instrumental variable $z$ which must be exogenous, correlated with the policy, and (directly) uncorrelated with the reward. In applications, the availability of an instrument can be problematic. 
Similarly, the application of the DID estimator can be problematic as well, as it requires longitudinal or repeated cross-sectional data. Not all the contexts can provide these types of data structures.  

\item \textit{Sensitivity analysis}. Sensitivity analysis can be used to assess the impact of unmeasured confounding variables on the decision carried out. By conducting a range of analyses that vary the assumptions about the strength of unmeasured confounding, sensitivity analysis can help to identify how robust the decision process is to potential violations of weak unconfoundedness.

\item \textit{Prior knowledge}. Prior knowledge about the relationship between the decision and the reward may be useful in identifying potential confounders that were not measured. This can help to reduce the impact of unmeasured confounding on the mapping between the decision and the reward.

\item \textit{Sensible assumptions}. Finally, sensible assumptions about the nature of unmeasured confounding can be used to develop statistical models that account for these confounding factors. For example, assuming that the unmeasured confounding variables have a similar effect on all actions can be used to adjust for their impact on the reward.
\end{itemize}

\section{Conclusions}
In data-driven optimal policy learning (OPL) with finite alternatives, the goal is to select the best alternative from a set of possible options based on a set of environmental inputs. This setting can be embedded within the family of \textit{contextual} multi-armed bandit models with observational data, where exploration was assumed to be already carried out and a large sample of past decisions, environmental features, and outcomes/rewards are available. Also, this may be seen as a simple but powerful framework used in data-driven reinforcement machine learning to select the optimal actions to undertake (optimal policy detection). 

Within this framework, this paper contributed in three directions by: (i) providing a brief review of the key approaches to estimating the reward (or value) function and optimal policy; (ii) delving into the analysis of decision risk and its consequences on optimal action detection; (ii) discussing the limitations/constrains of optimal data-driven decision-making by highlighting conditions under which optimal action detection can fail. 

The paper can be a valuable contribution by offering a concise yet thorough review of key approaches to estimating the reward (or value) function and optimal policy within the multi-action decision framework. By summarizing and analyzing these approaches, it can serve as a resource for researchers, practitioners, and decision-makers seeking an understanding of the current landscape of estimation methodologies for optimal decision.

By delving into the realm of decision risk within the given framework, the paper provides practical insights that can inform decision strategies in various domains. This analysis contributes to bridging the gap between theoretical concepts and their real-world applications, where decision-makers may have differential attitudes towards risk.

Finally, by discussing the limitations and constraints associated with optimal action detection, the paper adds a layer of realism to the effective use of OPL. This is crucial for guiding researchers and practitioners in understanding the conditions under which data-driven OPL may fall short, thereby paving the way for more nuanced and context-aware approaches.

\pagebreak

\pagebreak

\end{document}